\documentclass[conference]{IEEEtran}
\IEEEoverridecommandlockouts

\usepackage{cite}
\usepackage{tikz}
\usepackage{amsthm}
\usepackage{algorithm}
\usepackage{amsmath,amssymb,amsfonts}
\usepackage{algorithmic}
\usepackage{graphicx}
\usepackage{textcomp}
\usepackage{xcolor}
\def\BibTeX{{\rm B\kern-.05em{\sc i\kern-.025em b}\kern-.08em
    T\kern-.1667em\lower.7ex\hbox{E}\kern-.125emX}}
\usepackage{comment}

\usepackage{xspace} 
\usepackage{filecontents}

\usepackage{subcaption} 
\usepackage{booktabs}
\usepackage{multirow}

\newtheorem{definition}{Definition}

\newcommand{\Call}[2]{\textbf{Call}~#1(#2)}
\newcommand{\name}[0]{ROKA\xspace}

\begin{document}

\title{\Large \bf \name: Robust Knowledge Unlearning against Adversaries}

\author{\IEEEauthorblockN{
    Jinmyeong Shin\IEEEauthorrefmark{1},
    Joshua Tapia\IEEEauthorrefmark{1},
    Nicholas Ferreira\IEEEauthorrefmark{2},
    Gabriel Diaz\IEEEauthorrefmark{1},
    Moayed Daneshyari\IEEEauthorrefmark{2},
    Hyeran Jeon\IEEEauthorrefmark{1}
}\\
  \IEEEauthorblockN{
  \IEEEauthorrefmark{1}University of California, Merced, USA \\
  \IEEEauthorrefmark{2}California State University, East Bay, USA}
}

\maketitle

\begin{abstract}
The need for machine unlearning is critical for data privacy, yet existing methods often cause Knowledge Contamination by unintentionally damaging related knowledge. Such a degraded model performance after unlearning has been recently leveraged for new inference and backdoor attacks. Most studies design adversarial unlearning requests that require poisoning or duplicating training data. In this study, we introduce a new unlearning-induced attack model, namely \textit{indirect unlearning attack}, which does not require data manipulation, but exploits the consequence of knowledge contamination to perturb the model accuracy on security-critical predictions. 
To mitigate this attack, 
we introduce a theoretical framework that models neural networks as Neural Knowledge Systems. Based on this, we propose \textit{\name}, a robust unlearning strategy centered on \textit{Neural Healing}. Unlike conventional unlearning methods that only destroy information, \name\ constructively re-balances the model by nullifying the influence of forgotten data while strengthening its conceptual neighbors. 
To the best of our knowledge, our work is the first to provide a theoretical guarantee for knowledge preservation during unlearning. 
Evaluations on various large models, including vision transformers, multi-modal models, and large language models, 
show that \name\ effectively unlearns targets while preserving, or even enhancing, the accuracy of retained data, thereby mitigating the indirect unlearning attacks. 
\end{abstract}


\section{Introduction } \label{sec:intro}

\begin{figure*}[ht]
    \centering
    \begin{subfigure}[h]{0.3\textwidth}
        \centering
        \begin{subfigure}[h]{\textwidth}
            \centering
            \includegraphics[height=3cm]{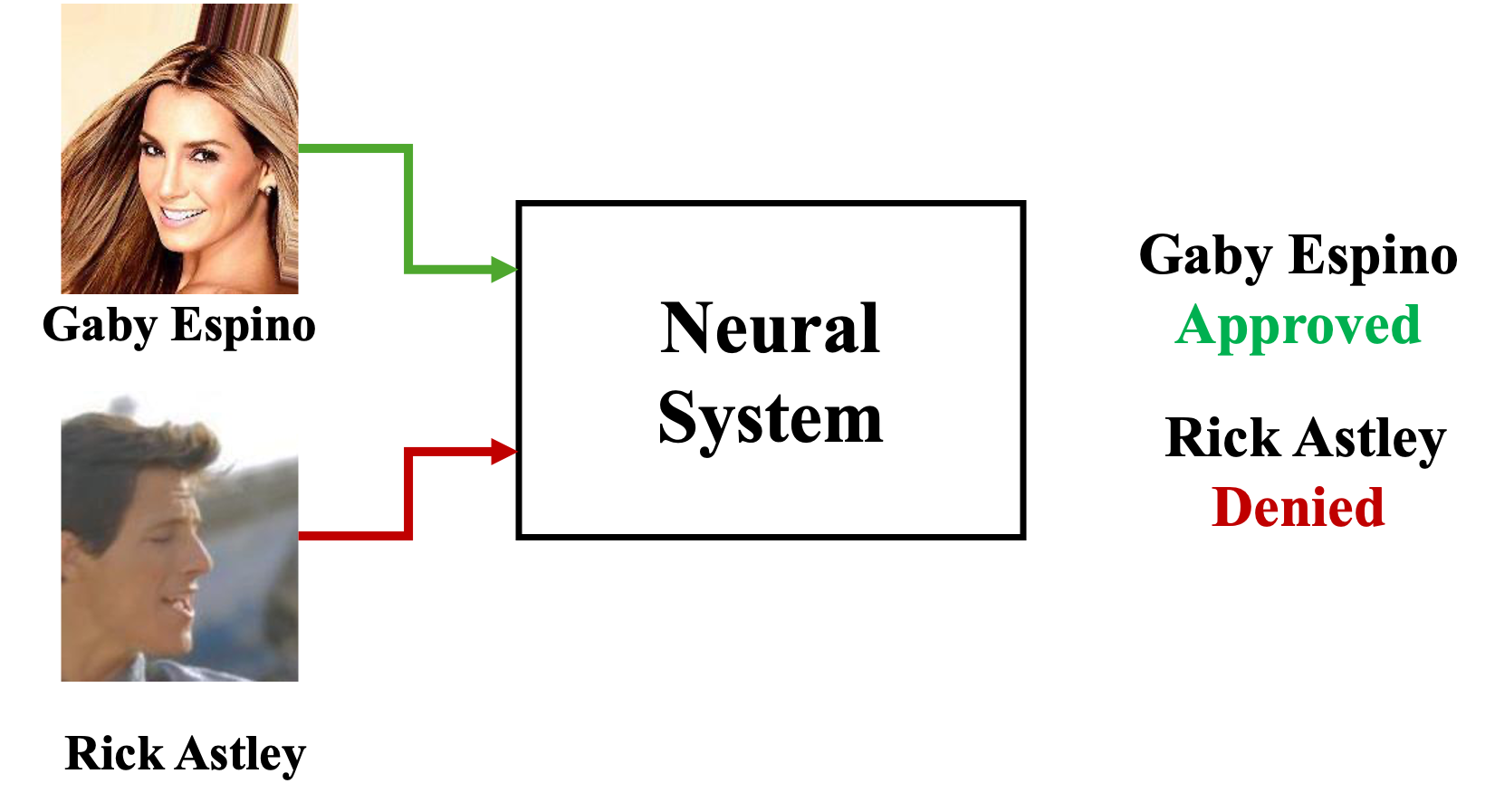}
        \end{subfigure}\vspace{-10pt}
        
        \caption{Original}
        \label{fig:threat_model:airplane}
    \end{subfigure}
    \begin{subfigure}[h]{0.3\textwidth}
        \centering
        \begin{subfigure}[h]{\textwidth}
            \centering
            \includegraphics[height=3cm]{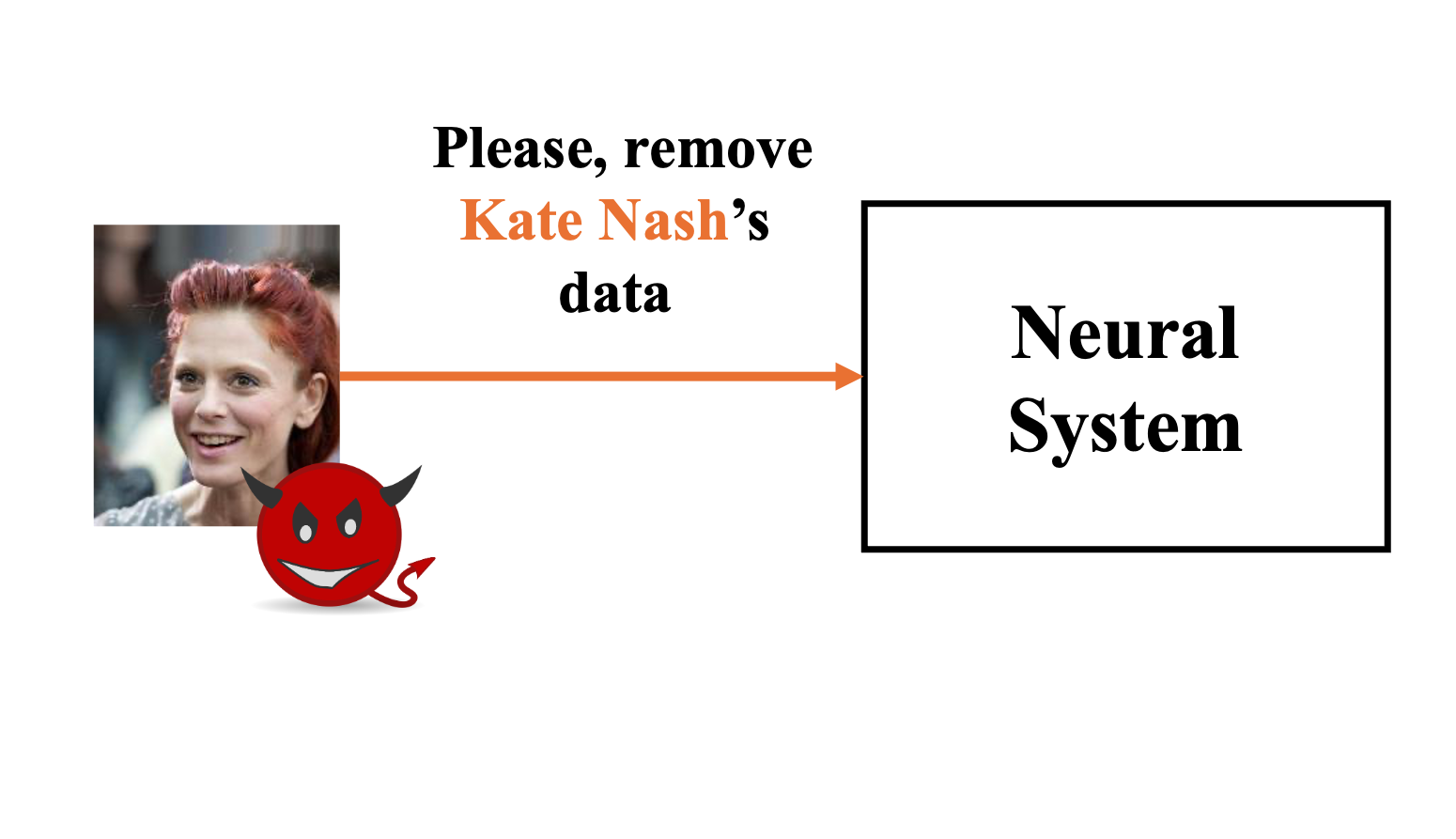}
        \end{subfigure}\vspace{-10pt}
        
        \caption{Adversarial Unlearning Request}
        \label{fig:threat_model:airplane}
    \end{subfigure}
    \begin{subfigure}[h]{0.3\textwidth}
        \centering
        \begin{subfigure}[h]{\textwidth}
            \centering
            \includegraphics[height=3cm]{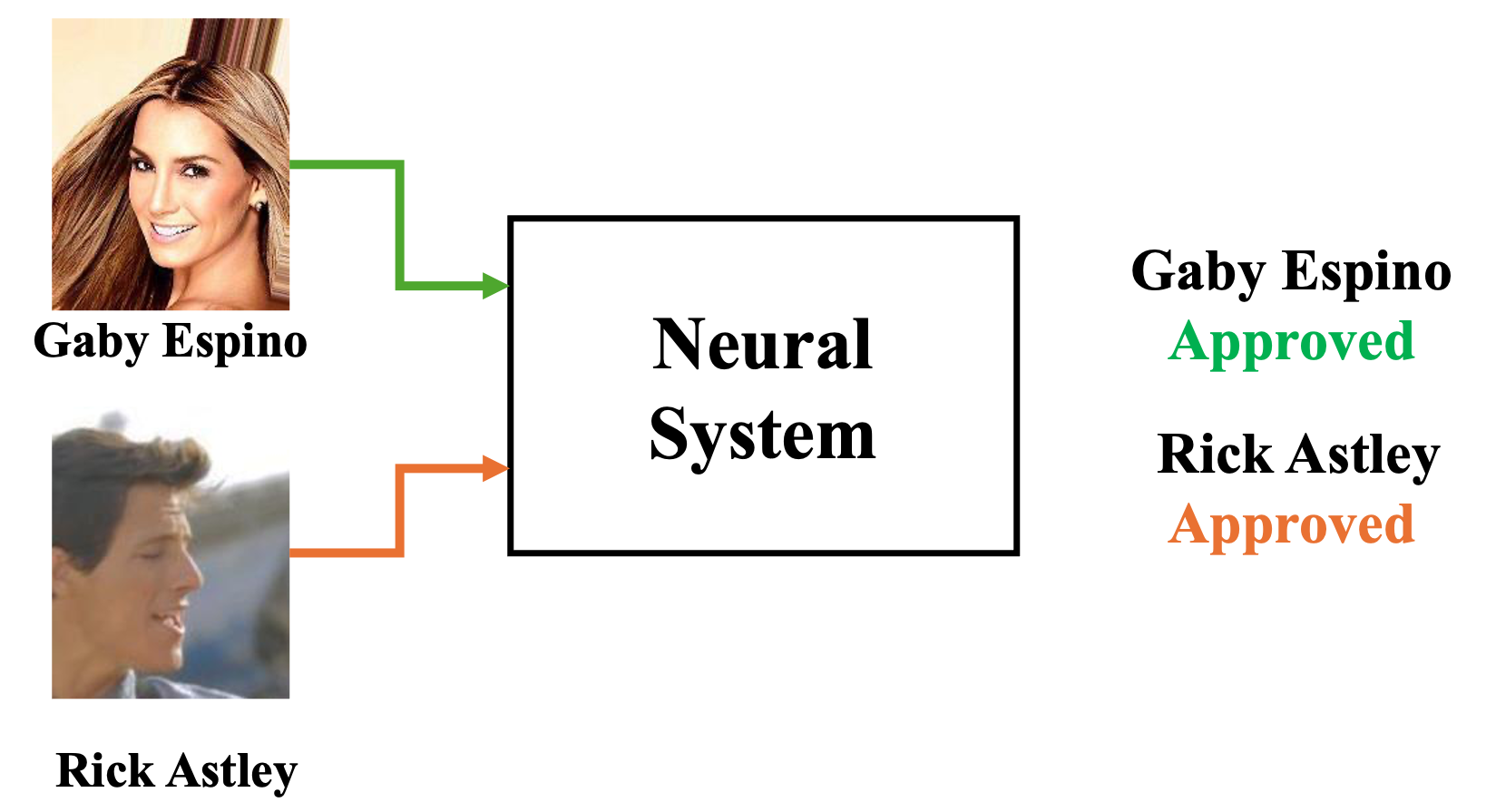}
        \end{subfigure}\vspace{-10pt}
        \
        \caption{After Unlearning}
        \label{fig:threat_model}
    \end{subfigure}
    \caption{Illustration of our proposed Indirect Unlearning Attack. This figure demonstrates how an unlearning request for one subject can compromise the security of another. (a) Initially, a hypothetical face recognition system correctly approves an authorized user (Gaby Espino) while denying an unauthorized one (Rick Astley). (b) An adversary then submits a request to unlearn a seemingly unrelated individual (Kate Nash). (c) After a conventional unlearning process, the model's knowledge is contaminated while it still correctly approves Gaby Espino, its ability to recognize Rick Astley is degraded, causing it to incorrectly grant him access and thus compromising the system's security.}\vspace{-15pt}
    \label{fig:threat_model}
\end{figure*}

The proliferation of large-scale machine learning~(ML) models, trained on vast datasets, has led to unprecedented capabilities across numerous domains. However, the abundant 
data that fuels these successes presents a significant and growing challenge: the need to forget. The ability to selectively remove specific information from a trained ML model, a process known as machine unlearning, is no longer a niche feature 
but a critical capability for responsible and ethical AI development. Recently, several data privacy policies have been developed, such as the General Data Protection Regulation~(GDPR)~\cite{gdpr-2011} and the California Consumer Privacy Act (CCPA)~\cite{ccpa-2018}, which grant individuals the right to be forgotten, mandating that companies be able to erase user data from their systems. 

With the growing importance of machine unlearning, several recent studies raised its security concerns~\cite{ye-2025-security, song-2025-security, huang-2024-iclr, zhao-2023-neurips, di-2023-neurips, huang-2024-security, liul-2024-aaai, naderloui-2025-security, pawelczyki-2025-iclr}. These studies demonstrated adversarial scenarios where an attacker requests the model owners to remove camouflaged or duplicated data with the intention to degrade the overall model's accuracy or trigger a backdoor after unlearning. Most of the attack models exploit the negative influences among neurons during the unlearning, which is particularly observed in inexact machine unlearning algorithms. 

Inexact machine unlearning has been introduced to address the computational and performance overhead of exact unlearning. Exact machine unlearning algorithms typically require retraining of the model with a modified training dataset excluding the data to forget. As the ML model size increases rapidly, where large language models (LLMs) use several billions to trillions of parameters, retraining-based exact unlearning becomes less practical, if not impossible. While several optimizations have been introduced, such as SISA 
training~\cite{9519428}, which retrains only the specific sub-model exposed to the data in question, such optimizations require a specialized and constrained training pipeline from the outset. Unlike exact unlearning, inexact unlearning uses a post-hoc approach; data are removed from a trained model. The representative inexact unlearning algorithms are \textit{gradient ascent unlearning}~(GA), which maximizes the loss on data to be forgotten; \textit{selective synaptic dampening},  identifies and weakens the parameters most critical to the forgotten knowledge using the Fisher Information Matrix; and \textit{selective unlearning (SEUL)}, which refines the GA process for LLMs by targeting specific spans of text.

Despite their efficiency, these existing inexact techniques share a significant risk of \textit{Knowledge Contamination}. This occurs when the process of removing unwanted information unintentionally damages related, desirable knowledge, degrading overall model performance. Gradient-based methods are imprecise, as a single update affects numerous parameters shared between forgotten and retained knowledge. This collateral damage is not just a random degradation of performance. Also, it creates a new attack surface. As we demonstrate in this paper, this effect can be exploited in an \textit{Indirect Unlearning Attack}, where an adversary intentionally requests the removal of one piece of information to strategically degrade a model's performance on another, security-critical task as illustrated in Figure~\ref{fig:threat_model}.

To address these challenges of \textit{Knowledge Contamination} and the consequential vulnerabilities, we introduce a new theoretical framework that conceptualizes neural networks as \textit{Neural Knowledge System}. Based on this framework, we propose \name, a robust unlearning strategy centered on \textit{Neural Healing}.  Unlike the conventional unlearning methods that only destroy information, \name constructively rebalances the knowledge contributions while unlearning. 
Through a novel \textit{Contribution Re-allocation} procedure, \name\ nullifies the influence of the forgotten data while proactively reinforcing its conceptual neighbors, thereby mitigating indirect unlearning. 

We demonstrate the effectiveness of ROKA through extensive evaluations on vision transformers (ViT~\cite{dosovitskiy2021imageworth16x16words}, DeiT~\cite{touvron2021training}), multi-modal models (CLIP~\cite{radford2021learningtransferablevisualmodels}), and LLMs (Llama 3.2~\cite{grattafiori2024llama3herdmodels}), tested on standard benchmarks including CIFAR~\cite{Krizhevsky09learningmultiple}, Tiny-ImageNet~\cite{Le2015TinyIV}, and MMLU~\cite{hendrycks2021measuringmassivemultitasklanguage}. The results show that \name effectively unlearns target data while preserving, and in some cases even enhancing, the accuracy of retained data.

The contributions of this paper are as follows:
\begin{itemize}
    \item We introduce a novel theoretical framework, the \textit{Neural Knowledge System}, to formalize knowledge representation in neural networks. Based on this, we provide the first theoretical guarantee for knowledge preservation during unlearning.
    \item We identify and empirically demonstrate a new vulnerability, the \textit{Indirect Unlearning Attack}, where conventional unlearning of one class can be exploited to strategically degrade a model’s accuracy on a different, security-critical class.
    \item We propose \name, a robust unlearning method centered on \textit{Neural Healing}. Through extensive evaluations on large-scale models like ViT, CLIP, and Llama, we show that \name effectively unlearns target data while preserving performance on retained data, offering a practical defense against the indirect unlearning attack. 
\end{itemize}

\section{Backgrounds} \label{sec:pre}

\subsection{Machine Unlearning }
Machine unlearning is the process of removing specific information from a trained ML model, a critical capability for data privacy, compliance with regulations like GDPR, and removing outdated or harmful data. The primary challenge is to forget the targeted information effectively while preserving the model's performance on all remaining data. While retraining a model from scratch on the data, excluding the targeted data to unlearn, 
provides a perfect unlearning outcome, retraining is often undesirable or even impractical due to the long training time and computational requirements. 
Consequently, research has focused on developing more efficient, approximate (i.e., inexact) unlearning methods. 

Bourtoule et al. propose \textit{sharded, isolated, sliced, and aggregated (SISA) training}, a framework that restructures the training pipeline to enable efficient, provably correct unlearning~\cite{9519428}. The data is partitioned into disjoint shards, and an ensemble of models is trained in isolation, one on each shard. To further reduce overhead, model parameters are checkpointed as data is introduced in incremental slices. An unlearning request only necessitates retraining a single, smaller model from an intermediate checkpoint. This approach significantly reduces the computational cost compared to retraining from scratch.

Jang et al. introduce a post hoc method for LLMs that performs \textit{gradient ascent  (GA) unlearning} on the token sequences to be forgotten~\cite{10.1609/aaai.v38i11.29092}. This approach reverses the standard training objective for the target data, effectively maximizing its loss. A key contribution is the finding that unlearning data sequentially in smaller batches is more stable and better preserves the model's general capabilities than unlearning a large set of data at once.

Foster et al. propose \textit{selective synaptic dampening~(SSD)}, a fast, retrain-free, and post hoc method~\cite{jang-etal-2023-knowledge}. SSD uses the Fisher Information Matrix (FIM) to identify parameters that are disproportionately important to the forget set. Instead of removing these parameters, it dampens their magnitude in proportion to their relative importance. This surgical modification targets specialized, memorized information and allows SSD to achieve performance competitive with retraining-based methods while modifying only a small subset of the model's weights.

Wang et al. introduce \textit{selective unlearning~(SEUL)}, a fine-grained method for LLMs that targets specific spans of text rather than entire data instances~\cite{Wang_Zeng_Guo_Wong_Gottlob_2025}. The unlearning objective maximizes the loss only on the tokens within the sensitive spans. This targeted approach is designed to minimize collateral damage to the model's general knowledge and better preserve its generative capabilities.

Given its popularity, we use gradient-ascent-based unlearning as a baseline in our evaluations.

\subsection{Unlearning-induced Attack Models}
While demands for machine unlearning increase due to the rising privacy concerns, several studies have raised security issues in machine unlearning. Some studies~\cite{ye-2025-security, song-2025-security, huang-2024-iclr, zhao-2023-neurips} demonstrated adversarial machine unlearning methods where an attacker intentionally request unlearning on data that cannot be unlearned due to duplicated copies in the training dataset, crafted data that do not exist in the training data with an intention of degrading the overall model-wise accuracy, and data that are critical for the model to reject harmful requests to compromise the safety of the target LLM. Another group of studies introduced backdoor and poisoning attacks activated by unlearning~\cite{di-2023-neurips, huang-2024-security, liul-2024-aaai}, where an attacker poisons a subset of the training dataset with camouflaged data to perturb model accuracy or trigger a backdoor after unlearning. To mitigate these models, recent studies suggest improving the evaluation metric of unlearning~\cite{naderloui-2025-security, pawelczyki-2025-iclr}, by using per-sample verifications with shadow models that can provide the expected model outputs after correct unlearning (e.g., unlearning via retraining) rather than using average model-wise accuracy, and applying Gaussian noise to clean data to understand and offset the impact of data poisoning attacks. 

Unlike these earlier studies, we introduce a new unlearning-induced attack model that does not require poisoning data. We leverage the diversity of unlearning influence among different data classes to indirectly compromise security-critical classes. We also introduce a mitigation method that uses a novel \textit{neuron healing}, which is much lightweight than earlier mitigations that require multiple shadow models.

\subsection{Layer-wise Relevance Propagation}\label{sec:lrp}
To design a robust knowledge unlearning, we leverage layer-wise relevance propagation (LRP) (Section~\ref{sec:healing}). 
LRP is an explainability technique used to attribute a model's prediction to its input features~\cite{bach2015pixel}. It operates by propagating the model's output score backward through the network's layers until it reaches the input, assigning a relevance score to each feature or neuron. The key idea of LRP is defining relevance between previous and next neurons, and its formulation into numbers that can be calculated. For a neuron $\mathbf{j}$ in a lower layer and a neuron $\mathbf{k}$ in higher layer, the relevance passed from $\mathbf{k}$ to $\mathbf{j}$ proportional to the contribution of $\mathbf{j}$ to $\mathbf{k}$'s activation. The general form of the relevance update rule is as follows:
$$R_j = \sum_k \frac{a_jw_{jk}}{\sum_i a_iw_{ik}} \cdot R_k$$
Here, $a_j$ is the activation of neuron $\mathbf{j}$, $w_{jk}$ is the weight connecting $\mathbf{j}$ to $\mathbf{k}$, and $R_k$ is the relevance of neuron $\mathbf{k}$. The denominator sums the contributions of all lower-level neurons to neuron $\mathbf{k}$, ensuring that the total relevance $R_k$ is conserved.

The $Input \cdot Gradient$ method~\cite{montavon2017explaining} is a simpler attribution technique where the importance of an input feature is calculated by multiplying the gradient of the output with respect to that feature by the feature's own value as follows:
$$R_{x_i} = x_i \cdot \frac{\partial f(\mathbf{x})}{\partial x_i}$$
It has been shown that if the basic LRP rule is applied uniformly across all layers of a deep neural network, the resulting relevance scores at the input layer are equivalent to the values computed by the $Input \cdot Gradient$ method.

\section{Indirect Unlearning Attack} \label{sec:threat}
We propose a new unlearning-induced attack model, namely \textit{indirect unlearning attack}, which unlearns a data class from a model with the intention of compromising the prediction accuracy of another security-critical data class.
Unlike an earlier study~\cite{huang-2024-iclr} that leverages the overall model-wise accuracy drop after unlearning, we exploit the imbalanced accuracy influence among data classes during unlearning. 

\begin{figure*}[ht]
    \centering
    \begin{subfigure}[h]{0.3\textwidth}
        \centering
        \begin{subfigure}[h]{\textwidth}
            \centering
            \includegraphics[width=\textwidth]{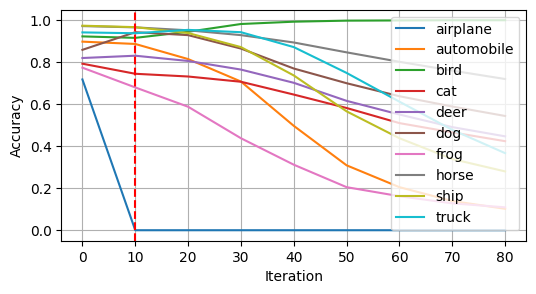}
        \end{subfigure}
        \begin{subfigure}[h]{\textwidth}
            \centering
            \includegraphics[width=\textwidth]{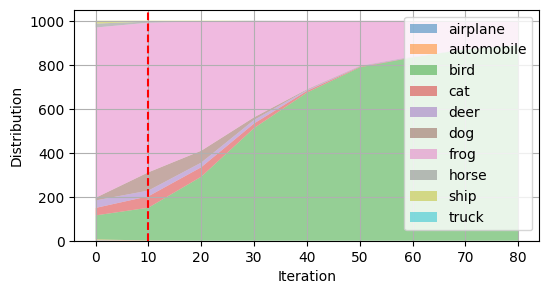}
        \end{subfigure}
        \caption{Airplane($C_{unlean}^{adv}$) - Frog($C_{target}^{adv}$)}
        \label{fig:threat_model:airplane}
    \end{subfigure}
    \begin{subfigure}[h]{0.3\textwidth}
        \centering
        \begin{subfigure}[h]{\textwidth}
            \centering
            \includegraphics[width=\textwidth]{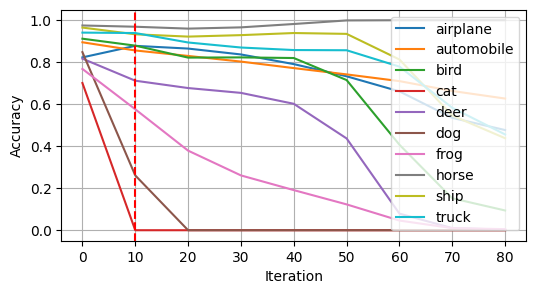}
        \end{subfigure}
        \begin{subfigure}[h]{\textwidth}
            \centering
            \includegraphics[width=\textwidth]{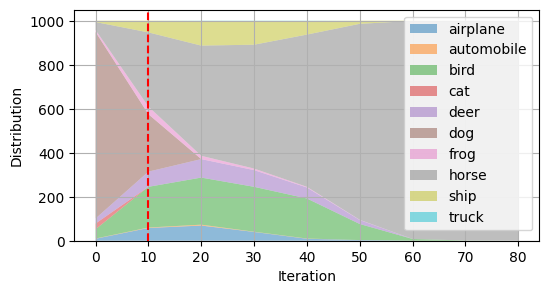}
        \end{subfigure}
        \caption{Cat($C_{unlean}^{adv}$) - Dog($C_{target}^{adv}$)}
        \label{fig:threat_model:airplane}
    \end{subfigure}
    \begin{subfigure}[h]{0.3\textwidth}
        \centering
        \begin{subfigure}[h]{\textwidth}
            \centering
            \includegraphics[width=\textwidth]{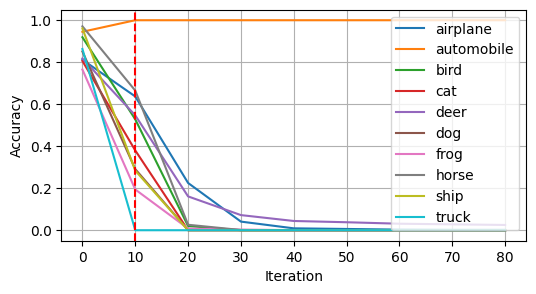}
        \end{subfigure}
        \begin{subfigure}[h]{\textwidth}
            \centering
            \includegraphics[width=\textwidth]{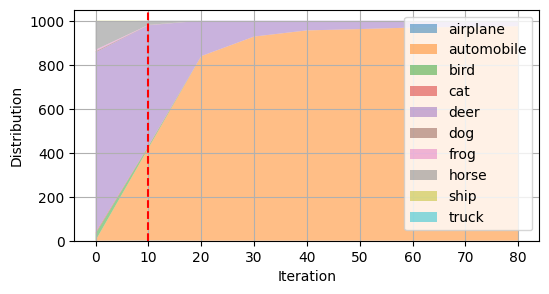}
        \end{subfigure}
        \caption{Truck($C_{unlean}^{adv}$) - Deer($C_{target}^{adv}$)}
        \label{fig:threat_model:airplane}
    \end{subfigure}
    \caption{Imbalanced prediction impacts from unlearning: $C_{unlean}^{adv}$ is the unlearned data class, $C_{target}^{adv}$ is the data class that exhibits the most severe accuracy drop among those 
    that were not unlearned. Our proposed Indirect unlearning attack aims to request to unlearn $C_{unlean}^{adv}$ to compromise the accuracy of $C_{target}^{adv}$, where $C_{target}^{adv}$ is assume to be a security-critical class.}\vspace{-15pt}
    \label{fig:indirect-unlearning}
\end{figure*}

\subsection{Threat Model }
\textbf{Unlearning Scenario}
According to GDPR, we assume that the attacker can raise a privacy concern and request the target model's owner to remove a specific class of data from the prediction. Then, the model owner uses a conventional inexact unlearning algorithm (e.g., GA-based unlearning) to make the model forget the class. We assume that retraining-based unlearning is practically undesirable due to the significant time and computing resources required to retrain the model.

\textbf{Attacker's Goal}
The attacker aims to degrade the prediction accuracy of a security-critical target class ($C^{adv}_{target}$) by intentionally requesting the model owner to unlearn another class ($C^{adv}_{unlearn}$). 
The attacker identifies $C^{adv}_{unlearn}$, which can negatively impact the prediction accuracy of $C^{adv}_{target}$ when $C^{adv}_{unlearn}$ is unlearned from the target model via conventional inexact unlearning methods. 
With the compromised model performance, the attacker aims to gain privileged access to the target system. For example, suppose that the attacker knows that a model's person recognition accuracy degrades significantly when a certain person's face data is unlearned. Then, the attacker intentionally requests to remove the person's face from the model prediction (claiming a privacy violation) to compromise the face recognition performance. If the model is used for a home authentication system, the degraded accuracy will lead to various security problems (e.g., false home intrusion detection and inability to unlock gates by the homeowner). 

\textbf{Attacker's Capabilities}
The attacker is assumed to have access to a subset of the training data, $D_{train}$, and be aware of the unlearning algorithm. We assume a gray-box attack. While the attacker does not have access to the model weight files, the attacker is aware of the target model architecture. The attacker uses random models that use the same architecture as the target model and are trained with the subset of $D_{train}$ that the attacker has access to. The attacker uses the models to test the accuracy influence between the subset classes in $D_{train}$.

\subsection{Indirect Unlearning \& Attack Model}\label{sec:threat:indirect}
We define the accuracy drop of the untargeted classes from unlearning as \textit{indirect unlearning}. While several studies reported overall model-wise accuracy drop after unlearning, we observe that there is one (or several) data classes that exhibit a notably severe accuracy degradation, even when those are not the target to forget. 

\subsubsection{Methodology}
To measure such accuracy influences among classes, we ran \textcolor{black}{experiments using the CIFAR-10 datasets and a pre-trained CLIP model (\texttt{clip-vit-base-patch32}). We performed a conventional GA
-based unlearning method. We investigated the indirect unlearning impacts by unlearning one of the classes (e.g., 10 in CIFAR-10) each time. 
We regularly measured the prediction accuracy of 
all classes to track the collateral impact of the unlearning process. By repeating this for different target classes, we identified the classes that are most negatively impacted by each unlearning target class. 
}

\subsubsection{Imbalanced Prediction Impact from Unlearning}\label{sec:threat:imbalance}
Figure~\ref{fig:indirect-unlearning} shows the per-class accuracy results during unlearning, when different classes are targeted to forget. The class to unlearn is denoted as $C^{adv}_{unlearn}$. As the top three graphs illustrate, unlearning affects the prediction performance of almost all data classes. The model owner would want to maintain the model’s overall prediction performance after unlearning. Therefore, we assume that the unlearning will stop early when the target class is nearly forgotten; we marked the point with a vertical red dotted line. 

Across the results, we were able to observe that the accuracy impact of the unlearning varies across the classes. There are severely impacted classes, such as \texttt{dog} having almost 60\% drop in (b) and \texttt{frog}, \texttt{ship}, \texttt{cat}, \texttt{bird}, and \texttt{deer} having 25\% to 58\% accuracy drop in (c). On the other hand, some classes preserve or derive even a better accuracy after unlearning; \texttt{bird} in (a), \texttt{horse} in (b), and \texttt{automobile} in (c). These can be explained by the bottom three graphs, where each graph shows the prediction changes of the most impacted classes during the unlearning. As unlearning continues, \texttt{frog} becomes predicted as \texttt{bird} in (a), \texttt{dog} as \texttt{horse} in (b), and \texttt{deer} as \texttt{automobile} in (c). Thus, those classes' predictions are illusioned to be improved. Such an imbalanced accuracy impact was also observed under different unlearning algorithms. Table~\ref{tab:classification_ratio:GA} in Section~\ref{sec:eval} shows more precise and comprehensive cases of imbalanced prediction manner after unlearning.

\subsubsection{Indirect Unlearning Attack}
Our proposed indirect unlearning attack leverages the imbalanced accuracy impact from unlearning. As assumed in the threat model, the attacker has one or multiple random models that use the same architecture as the target model. From the subset of the training dataset ($D_{train}$), the attacker wants to degrade the prediction accuracy of a target security-critical class, $C^{adv}_{target}$. 

First, the attacker trains the random models with the subset of $D_{train}$. Then, the attacker iterates the classes in the subset of $D_{train}$, unlearns each class each time, and checks the accuracy of $C^{adv}_{target}$ after the unlearning. The unlearning stops when the unlearning class's accuracy becomes almost zero. From these iterations of unlearning, the attacker identifies ($C^{adv}_{unlearn}$) that severely drops the accuracy of the actual attack target class, ($C^{adv}_{target}$). The attacker uses the same unlearning algorithm as the model owner would use, as assumed in the threat model. To make the attack effective, the attacker only considers the classes that drop the accuracy of $C^{adv}_{target}$ by more than a predefined threshold, $\tau^{adv}_{target}$. We used the threshold 25\% in our experiments. For example, in Table~\ref{tab:classification_ratio:GA}, the classes highlighted with red font are those experiencing more than 25\% accuracy decrease when the corresponding target class in the same row is unlearned. Thus, one of the red-colored classes can be $C^{adv}_{target}$ and the corresponding target class can be selected as $C^{adv}_{unlearn}$. Then, the attacker requests the model owner to unlearn the identified $C^{adv}_{unlearn}$. After unlearning is completed, the attacker verifies its attack success by checking the prediction accuracy of $C^{adv}_{target}$ and querying the model with security-critical tasks (e.g., authentication).

\section{\name: Robust Knowledge Unlearning through \textit{Neural Healing}} \label{sec:roka}
To mitigate the proposed indirect unlearning attack, we propose a robust knowledge unlearning method, \name. ROKA tackles indirect unlearning by \textit{healing} the neurons that are unintentionally affected by the unlearning. With ROKA, unlearning can more safely remove only the target classes with limited impact on the other classes. To identify the \textit{damaged} neurons for healing, we propose to leverage LRP propagations.

We first formally analyze the knowledge destruction in modern unlearning algorithms by defining a neural network as a \textit{hierarchical knowledge system} (Section~\ref{sec:analysis}). Then, we describe \name in the context of the hierarchical knowledge system (Section~\ref{sec:healing}). 

\subsection{Analysis of Knowledge Destruction in Machine Unlearning}\label{sec:analysis} 

\subsubsection{Defining a Neural Knowledge System}

A neural knowledge system can be conceptualized as a transformative process that maps inputs to outputs through an intermediate, abstract representation space. We define this system $\mathcal{S}$ as follows:
\begin{equation}
    \mathcal{S} = (\mathbf{X}, \mathbf{K}, \mathcal{F})
\end{equation}
where $\mathbf{X}$ is the input space of raw data, $\mathbf{K}$ is the internal space of abstract knowledge representations, and $\mathcal{F}$ is a set of functions governing the transformations between these spaces including primary transformations 
such as \textit{Encode} \{$\mathcal{E}: \mathbf{X} \to \mathbf{K}$\} and \textit{Decode} \{$\mathcal{D}: \mathbf{K} \to \mathbf{Y} $\}, which are equivalent to pre-processing of inputs and post-processing of outputs in a neural network. 






\subsubsection{Hierarchical Representation of Neural Network}

The internal knowledge space $\mathbf{K}$ is not monolithic. It is structured as a hierarchy of levels $(\mathbf{K}_0, \mathbf{K}_1, ..., \mathbf{K}_L)$ where each level $\mathbf{K}_l$ represents a progressively higher degree of abstraction. This form is similar to Markov chain $\mathbf{K}_{l-1} \to \mathbf{K}_l$, where a state in $\mathbf{K}_l$ is an aggregation of states from $\mathbf{K}_{l-1}$. The parameters in later layers of a network are often highly specialized to specific features, which supports this hierarchical view of knowledge abstraction.



\subsubsection{Properties of Neural Knowledge}
\noindent \textbf{Static Property} Given the hierarchical structure, we can define static properties that characterize a knowledge set and its components at a fixed level of abstraction.


\begin{definition}
    \textbf{Contribution} ($C$) of a lower-level component $k_{l-1},i$ to a higher-level state $\mathbf{K}_{l,j}$ measures \textbf{its local importance within that specific aggregation}, $C(k_{l-1,i} \to \mathbf{K}_{l,j})=\frac{w_{l-1,j}}{\mathbf{W}_{l,j}}$. It is defined as its proportional weight, which is equivalent to the conditional probability $p(k_{l-1,i}|\mathbf{K}_{l,j})$.
\end{definition}
This metric is important to our neural knowledge unlearning algorithm (Section~\ref{sec:healing}), 
which uses it to identify and re-balance the influence of specific knowledge components. 

\hspace{1em}

\noindent \textbf{Dynamic Properties} Dynamic properties emerge from the flow and transformation of information between the hierarchical levels of the neural knowledge system.
\begin{definition}
    \textbf{Influence} measures \textbf{the sensitivity of the entire system's entropy} to a change in a single component. It is equivalent to the information content of the destination group that the component belongs to:
\end{definition}\vspace{-10pt}
$$\text{Influence}(w_i \to \mathbf{K}_l) = -\log (W_{dest})$$

While individual influences vary, the \textit{Expected Influence} is a conserved quantity. The total average sensitivity is a fundamental property of the final knowledge state and is constant regardless of which preceding level is measured:
$$E[\text{Inf}(\mathbf{K}_{l-1} \to \mathbf{K}_l)]=E[\text{Inf}(\mathbf{K}_{l-2} \to \mathbf{K}_l)]$$
To reconcile the conserved average influence with the observation that low-level changes can have out-sized effects, \textit{Leverage} is defined.

\begin{definition}
    \textbf{Leverage} is influence scaled by the component's own weight. It quantifies the potential for a component to cause disproportionate change.
\end{definition}\vspace{-10pt}
$$\text{Leverage}(w_i \to \mathbf{K}_l)=\frac{\text{Influence}}{w_i}=\frac{-\log(W_{dest})}{w_i}$$
Leverage is highest for small, foundational components and decreases with aggregation: 
$$\text{Leverage}(\mathbf{K}_{l-2})>\text{Leverage}(\mathbf{K}_{l-1})$$
This property explains the fragility of complex systems to perturbations in their fundamental components, as a small absolute change to a low-weight component constitutes a large relative change, which can cascade through the hierarchy.

\subsubsection{Knowledge Destruction in Neural Knowledge System}
A catastrophic event, \textit{Knowledge Destruction}, can occur within the knowledge hierarchy where a minor change to a foundational component, amplified by high Leverage, causes a disproportionately large and incoherent shift in a higher-level knowledge representation, leading to unpredictable outputs.

\begin{definition}
    \textbf{Knowledge Destruction} is a state transition within the neural knowledge system characterized by a dramatic change in a high-level knowledge set~($\mathbf{K}_l$) caused by a small perturbation of a lower-level component~($k_{l-1,i}$).
\end{definition}
This effect is most pronounced when the perturbed component has high \textit{Leverage}.

The resulting higher-level representation, $\mathbf{K'}_l$, becomes inconsistent with its neighboring states, causing the system's \textit{Decode}~$\mathcal{D}$ to produce outputs that are unstable and radically different from what would be expected. It is a form of induced catastrophic forgetting, not for an entire task, but for a localized region of the model's knowledge space.

\subsubsection{The Boundary of Knowledge Destruction}
The boundary of Knowledge Destruction is the threshold at which the change in the system's final output becomes significant and unpredictable. We can define this boundary by measuring the dissimilarity between the output distribution before and after the perturbation using the \textit{Kullback-Leibler~(KL) Divergence}~\cite{Kullback_1951}.

\textit{Knowledge Destruction} occurs when the KL Divergence between the original and perturbed output distributions exceeds a critical threshold, $\tau$. This output change is driven by the relative change in the low-level component's weight, amplified by its \textit{Leverage}:
$$D_{KL}(P(\mathbf{Y}|\mathbf{K'}_l) \ || \ P(\mathbf{Y}|\mathbf{K}_l)) \ \propto \ \ \ \ \ \ \ \ \ \ \ \ $$ 
$$\ \ \ \ \ \ \ \ \ \ \ \text{Leverage}(w_{l-1,i} \to \mathbf{K}_l) \cdot \left| \frac{\Delta w_{l-1,i}}{w_{l-1,i}} \right|$$

Therefore, the boundary condition for \textit{Knowledge Destruction} is met when:
$$C_{p} \cdot \text{Leverage}(w_{l-1,i} \to K_l) \cdot \left| \frac{\Delta w_{l-1,i}}{w_{l-1,i}} \right| > \tau$$
where $D_{KL}(...)$ is the KL Divergence measuring the change in the final output, $\tau$  is a predefined threshold for what constitutes a dramatic change, $\text{Leverage}(...)$ is the leverage of the perturbed low-level component, $\left| \frac{\Delta w_{l-1,i}}{w_{l-1,i}} \right|$ is the relative magnitude of the perturbation, and $C_{p}$ is a system-dependent constant of proportionality. 

In essence, \textit{Knowledge Destruction} occurs when a relatively small change to a foundational component is amplified by high leverage to such a degree that it crosses a critical threshold, destabilizing the system's output.

\begin{figure*}[t]
    \centering
    \hfill
    \begin{subfigure}[h]{0.2\textwidth}
        \centering
        \includegraphics[height=3.8cm]{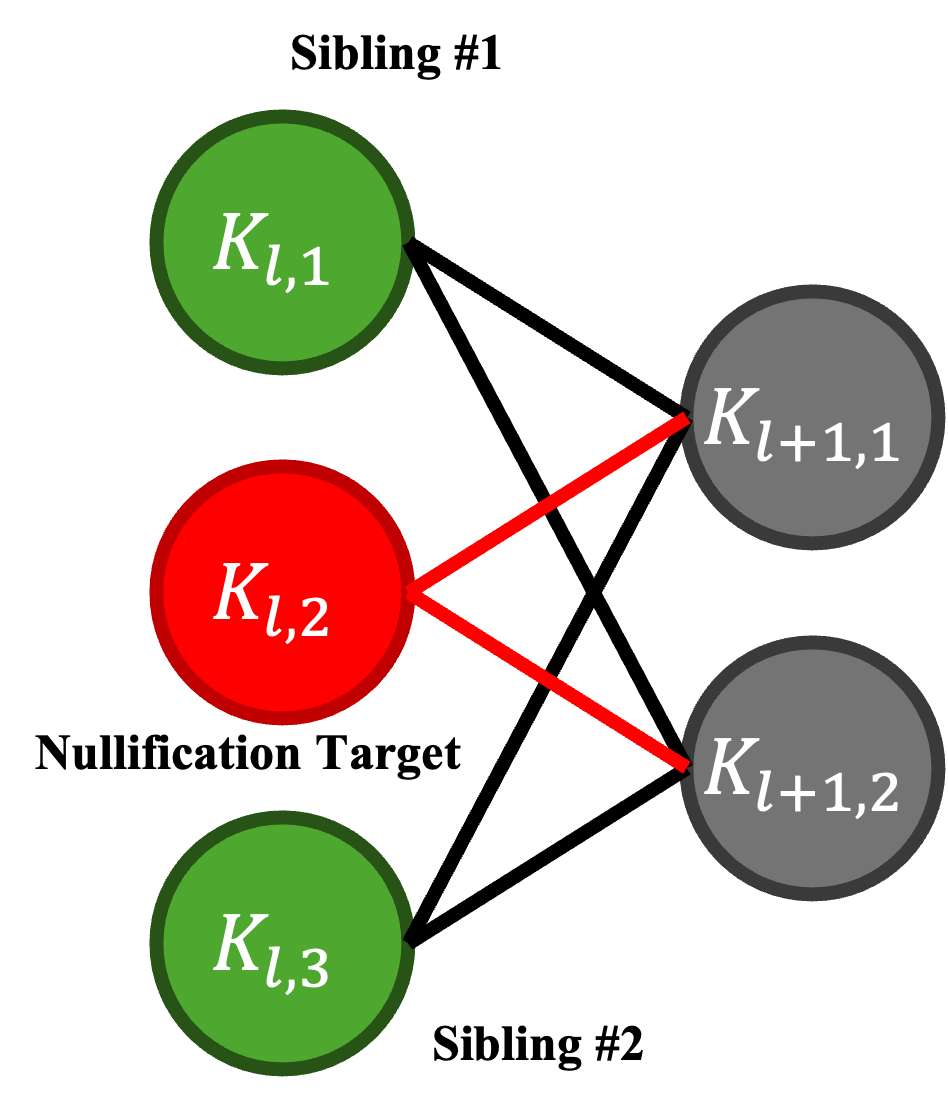}
        \subcaption{Target and Siblings}
    \end{subfigure}
    \hfill
    \begin{subfigure}[h]{0.65\textwidth}
        \centering
        \includegraphics[height=3.8cm]{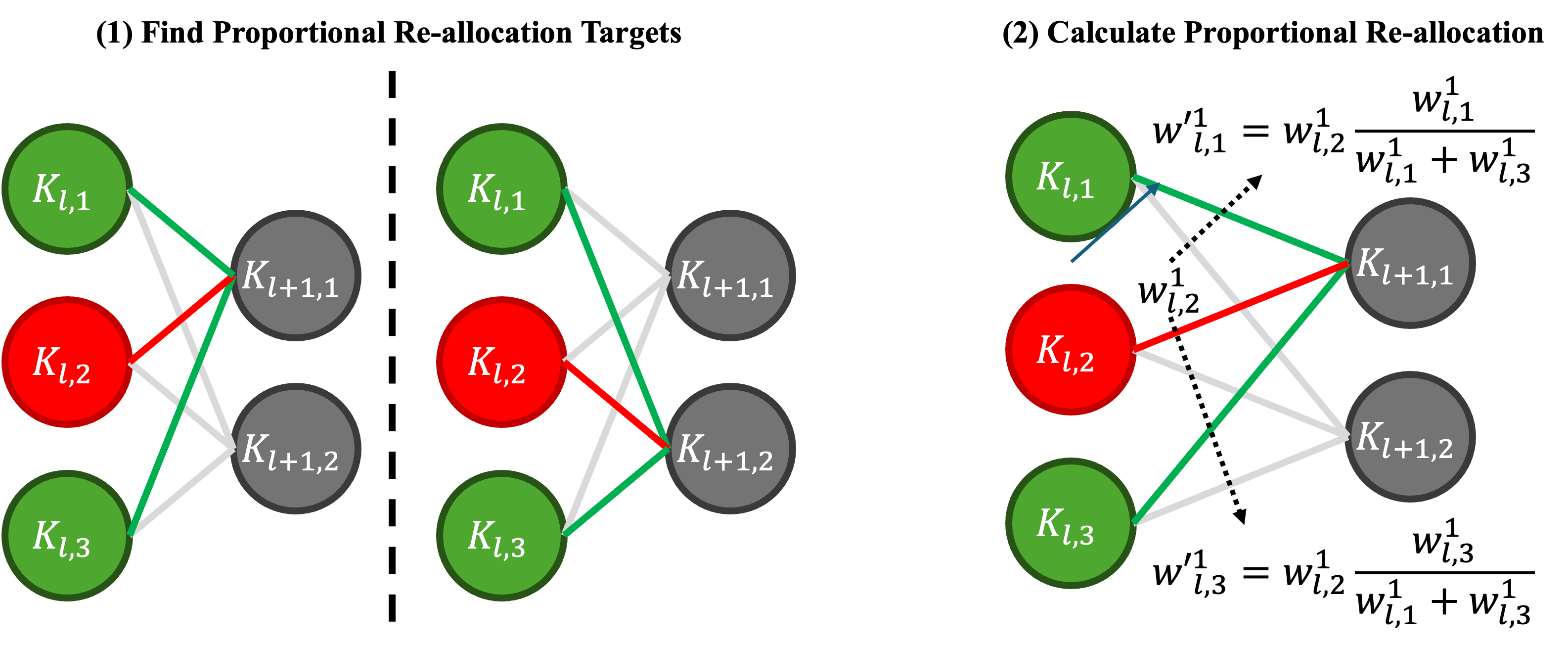}
        \subcaption{Proportional Re-allocation}
    \end{subfigure}
    \hfill
    \caption{Contribution Re-allocation Procedure: When unlearning a knowledge encoded in $K_{l,2}$, the nullified target information weakens the contribution of the next layer neurons, which results in an indirect unlearning of the knowledge delivered with sibling neurons. Our proposed \textit{Neural Healing} reallocates the original contribution of $K_{l,2}$ to the sibling neurons, $K_{l,1}$ and $K_{l,3}$, to preserve the total contribution, thereby the reallocated extra contribution factor can strengthen (maintain) their knowledge. }\vspace{-10pt} \label{fig:contribution-reallocation}
\end{figure*}

\subsubsection{The Limitations of Existing Unlearning}
The primary limitation of existing unlearning is the risk of unintentionally triggering \textit{Knowledge Destruction} in concepts that should be retained. We can define this collateral damage as \textit{Knowledge Contamination}.
\begin{definition}
\textbf{Knowledge Contamination} is the process where an unlearning operation targeted at a specific knowledge set, $\mathbf{K}_{forget}$, causes an unintended and significant increase in the Entropy of a related, retained knowledge set, $\mathbf{K}_{retain}$.
\end{definition}

\paragraph{Limitation of Gradient Ascent}
 The main drawback of gradient-based methods is their imprecision. A gradient update affects a wide array of parameters, many of which are shared between forgotten and retained knowledge. This process can easily cross the \textit{Boundary of Knowledge Destruction} for neighboring concepts, causing the scarring effect. This leads to a decrease in the model's overall \textit{Information Density} and a degradation in performance on retained data.



 We can formalize the boundary where this collateral damage occurs. \textit{Knowledge Contamination} happens when the unlearning update $\Delta w_{shared}$ applied to a shared parameter is large enough to cross the destruction threshold for the retained knowledge it supports:
 $$C \cdot \text{Leverage}(w_{shared} \to K_{retain}) \cdot \left| \frac{\Delta w_{shared}}{w_{shared}} \right| > \tau_{retain}$$
 This equation defines the fundamental challenge of unlearning: an update strong enough to forget one concept can contaminate another if they share influential, high-leverage components. The most advanced unlearning methods are those that can induce \textit{Knowledge Destruction} on a target while staying below the \textit{Knowledge Contamination boundary} for all other concepts.

\subsection{Neural Healing for Knowledge System}\label{sec:healing}

The task of machine unlearning, within the context of our theoretical framework, is to surgically remove a specific set of knowledge components, $\mathbf{K}_{forget}$, from a trained neural knowledge system. The ideal outcome is a final model state that is information-theoretically indistinguishable from a system that was never exposed to the forgotten data in the first place. This must be achieved while preserving the integrity and performance of the vast set of retained knowledge, $\mathbf{K}_{retain}$.

The primary challenge in this study is avoiding uncontrolled \textit{Knowledge Destruction}. As discussed, naive unlearning methods, such as those based purely on gradient ascent, can be effective at inducing destruction on the targeted knowledge but often at a great cost. Their imprecision frequently leads to \textit{Knowledge Contamination}, where the destructive process spills over and damages related, retained knowledge.  This results in a scarred system characterized by \textit{Influence Anomalies} and degraded \textit{Information Density}. To overcome these limitations, a more sophisticated approach is required, one that focuses not just on removal, but on actively healing the knowledge structure.

To this end, we propose \textit{Neural Healing} to reduce \textit{Knowledge Contamination}, thereby mitigating various attack models that leverage uncontrolled \textit{Knowledge Destruction}, such as our proposed indirect unlearning attack. The principle of \textit{Neural Healing} is a constructive approach to unlearning that theoretically guarantees a scar-free result. Instead of aggressively destroying unwanted information and leaving a structural void, this strategy re-balances the knowledge system by nullifying the influence of a forgotten component and strengthening its conceptual neighbors to compensate. This restores the system to a new, stable, and coherent state through a novel 
procedure named as \textit{Contribution Re-allocation}.

\subsubsection{Contribution Re-allocation Procedure}
The contribution re-allocation procedure is illustrated in Figure~\ref{fig:contribution-reallocation}. Given a component $k_{l-1,i}$ to be forgotten with weight $w_{l-1,i}$, the procedure is as follows:

\vspace{1em}

\noindent \textbf{Nullification}:
The contribution of $k_{l-1, i}$ is eliminated from the system. This creates a weight deficit equal to $w_{l-1, i}$ that must be re-distributed.

\noindent \textbf{Identification of Siblings}:
A sibling set, $\mathbf{S}_i$ , is identified. This set contains all components in the same hierarchical layer that are structurally related to $k_{l-1, i}$, typically by being part of the same parent aggregation in the subsequent layer $\textbf{K}_l$.

\noindent \textbf{Proportional Re-allocation}:
The weight deficit is re-allocated among the members of the sibling set $\mathbf{S}_i$. The weight added to each sibling $k_{l-1,j} \in \mathbf{S}_i$, denoted $\Delta w_j$, is proportional to that sibling's original \textit{Contribution} to the local group. This is calculated by first finding the total weight of the sibling set, $W_{siblings}$:
$$W_{siblings} = \sum_{k_{l-1, k} \in \mathbf{S}_i} w_{l-1, k}$$
The update $\Delta w_j$ applied to each sibling $k_{l-1, j}$ is then:
$$\Delta w_j = w_{l-1, i} \cdot \frac{w_{l-1, j}}{W_{siblings}}$$
The final weight of the healed sibling component becomes:
$$w'_{l-1, j} = w_{l-1, j} + \Delta w_j$$
This procedure ensures that the total weight of the parent group is conserved, maintaining the integrity of the knowledge hierarchy and preventing the \textit{Density Scarring} and \textit{Influence Anomalies} characteristic of destructive methods.

\subsubsection{Practical Limitations of Implementation}
While this procedure provides a theoretically ideal, scar-free 
unlearning, a strict and direct implementation in a modern neural network is computationally infeasible. The primary limitation is that knowledge in a neural network is highly distributed; the influence of a single data point is not represented by a single conceptual weight, $w_{l-1, i}$, but is instead entangled across millions of synaptic weights in a non-local manner. Therefore, it is challenging or even impossible to identify 
a single parameter to nullify. 

Approximating the \textit{Contribution} of a single data point to every parameter in the network requires computationally expensive attribution methods. Performing such a procedure for the forgotten point and all of its neighbors for every unlearning step would introduce significant computational overhead comparable to the cost of partial retraining, 
which these methods aim to avoid. 
Indeed, similar attribution-based unlearning methods have been found to be exceptionally slow in practice~\cite{wang2024attributebyunlearning}. Therefore, we propose a practical design to realize \textit{Contribution Re-allocation} by using efficient proxies, such as the \textit{Contribution-Weighted Loss}, in the following section. 

\subsection{Stochastic Unlearning with Neural Healing}
To bridge the gap between theory and practice, we introduce 
\textit{Stochastic Unlearning with Neural Healing}, 
which is a practical implementation of \textit{Neural Healing}. 
Instead of attempting a single, perfect surgical update, this method performs a series of small, iterative adjustments that stochastically guide the neural knowledge system toward the desired unlearned state. As proposed in \textit{Neural Healing}, this approach removes unwanted knowledge while proactively preserving and reinforcing other knowledge at the same time. 

We present two variations of this strategy tailored to different unlearning scenarios: (1)~one for when the knowledge to be forgotten is explicitly defined, and (2)~another for the more complex case where the unwanted knowledge is only implicitly defined by a set of unlabeled data.

\subsubsection{Targeted Stochastic Unlearning with Neural Healing}

\begin{algorithm}[t]
\caption{Targeted Stochastic Unlearning with Neural Healing}
\label{alg:targeted}
\begin{algorithmic}[1]
\REQUIRE Model $M$, Dataset $D$, Target Label $l_{\text{target}}$, Iterations $N_{\text{iter}}$, Healing Factor $\alpha$, Learning Rate $\eta$
\ENSURE Unlearned Model $M'$

    \STATE Initialize optimizer $\mathcal{O}$ for model $M$ with learning rate $\eta$
    \STATE Initialize data sampler $\mathcal{S}$ for dataset $D$

    \STATE
    
    \FOR{$i \leftarrow 1$ to $N_{\text{iter}}$}
        \STATE $B \leftarrow \Call{GetBatch}{\mathcal{S}}$
        
        \STATE $P_{\text{scores}} \leftarrow []$
        \FORALL{$x \in B$}
            \STATE $p \leftarrow M(x)[l_{\text{target}}]$
            \STATE Append $p$ to $P_{\text{scores}}$
        \ENDFOR
        \STATE $x_{\text{forget}} \leftarrow \Call{SampleProportionally}{B, P_{\text{scores}}}$
        \STATE
        \STATE $R \leftarrow \Call{CalculateContributions}{M, B}$ 
        \STATE $r_{\text{forget}} \leftarrow R[x_{\text{forget}}]$
        \STATE $B_{\text{siblings}} \leftarrow \Call{FindNearestNeighbors}{R, r_{\text{forget}}}$
        \STATE
        \STATE $\mathcal{L}_{\text{forget}} \leftarrow \Call{CrossEntropy}{M(x_{\text{forget}}), l_{\text{target}}}$
        \STATE $y_{\text{pred}} \leftarrow \arg \max M(B_{\text{siblings}})$
        \STATE $\mathcal{L}_{\text{heal}} \leftarrow \Call{CrossEntropy}{M(B_{\text{siblings}}), y_{\text{pred}}}$
        \STATE
        \STATE $\mathcal{L}_{\text{unlearn}} \leftarrow \mathcal{L}_{\text{forget}} - \alpha \cdot \mathcal{L}_{\text{heal}}$
        \STATE $\nabla \leftarrow \Call{ComputeGradients}{-\mathcal{L}_{\text{unlearn}}}$
        \STATE $\Call{UpdateParameters}{M, \mathcal{O}, \nabla}$
    \ENDFOR
    \STATE \RETURN $M$
\end{algorithmic}
\end{algorithm}

In the targeted unlearning scenario, the knowledge to be forgotten is explicitly defined by a target label, $l_{\text{target}}$. The objective is to reduce the model's ability to associate any data with this label while preserving its performance on all other concepts. The procedure, detailed in Algorithm~\ref{alg:targeted}, implements 
the principles of Neural Healing in an iterative, optimization-based framework.

The Algorithm~\ref{alg:targeted} begins by sampling a batch of data from the full dataset. Within this batch, it first identifies a single data point, $x_{\text{forget}}$, to serve as the focus of the unlearning step. This selection is stochastic and is guided by the model's own current predictions; data points for which the model has high confidence in the target label $l_{\text{target}}$ are more likely to be selected. This probabilistic approach ensures that the unlearning effort is continuously focused on the data most representative of the knowledge to be forgotten.

Once $x_{\text{forget}}$ is identified, the algorithm identifies its sibling knowledge. This is achieved by first calculating a contribution map for the entire batch, which attributes the model's output to its internal representations. This procedure is the same as the relevance calculation procedure of $Input \cdot Gradient$ method for LRP (Section~\ref{sec:lrp}). The siblings, $B_{\text{siblings}}$, are then identified as the nearest neighbors to $x_{\text{forget}}$ within this contribution space.

The important part of the method is the composite loss function. The forgetting loss, $\mathcal{L}_{\text{forget}}$, is the standard cross-entropy loss for $x_{\text{forget}}$  against the target label. The healing loss, $\mathcal{L}_{\text{heal}}$, is a self-distillation objective. It encourages the model to remain stable on the sibling data by minimizing the cross-entropy loss against its own, unperturbed predictions, $y_{\text{pred}}$.  The final unlearning loss combines these two objectives:
$$\mathcal{L}_{\text{unlearn}} = \mathcal{L}_{\text{forget}} - \alpha \cdot \mathcal{L}_{\text{heal}}$$

By performing gradient ascent on this composite loss, equivalent to gradient descent on $-\mathcal{L}_{\text{unlearn}}$, the optimizer is simultaneously instructed to maximize the error on the forgotten point while minimizing the error on its neighbors. This single update step effectively re-allocates the model's focus away from the unwanted knowledge and towards the sibling knowledge, executing a practical, computationally feasible step of \textit{Neural Healing}.

\subsubsection{Non-targeted Stochastic Unlearning with Neural Healing}

\begin{algorithm}[t!]
\caption{Non-Targeted Stochastic Unlearning with Neural Healing}
\label{alg:non-targeted}
\begin{algorithmic}[1]
\REQUIRE Model $M$, Forget Dataset $D_{\text{forget}}$, Iterations $N_{\text{iter}}$, Healing Factor $\alpha$, Learning Rate $\eta$
\ENSURE Unlearned Model $M'$

    \STATE $Y_{\text{pseudo}} \leftarrow \Call{PredictAll}{M, D_{\text{forget}}}$
    \STATE $R_{\text{all}} \leftarrow \Call{CalculateAllContributions}{M, D_{\text{forget}}}$
    \STATE $r_{\text{centroid}} \leftarrow \Call{Mean}{R_{\text{all}}}$
    \STATE $r_{\text{centroid}} \leftarrow r_{\text{centroid}} / \|r_{\text{centroid}}\|_2$
    \STATE
    \STATE Initialize optimizer $\mathcal{O}$ for model $M$ with learning rate $\eta$
    \STATE Initialize data sampler $\mathcal{S}$ for $D_{\text{forget}}$
    \STATE
    
    \FOR{$i \leftarrow 1$ to $N_{\text{iter}}$}
        \STATE $B, B_{\text{indices}} \leftarrow \Call{GetBatchWithIndices}{\mathcal{S}}$
        \STATE
        \STATE $R_{\text{current}} \leftarrow \Call{CalculateContributions}{M, B}$
        \STATE $R_{\text{current}} \leftarrow R_{\text{current}} / \|R_{\text{current}}\|_2$
        \STATE $s_{\text{scores}} \leftarrow \Call{CosineSimilarity}{R_{\text{current}}, r_{\text{centroid}}}$
        \STATE $idx_{\text{forget}} \leftarrow \arg \max(s_{\text{scores}})$
        \STATE $x_{\text{forget}} \leftarrow B[idx_{\text{forget}}]$
        \STATE
        \STATE $r_{\text{forget}} \leftarrow R_{\text{current}}[idx_{\text{forget}}]$
        \STATE $B_{\text{siblings}} \leftarrow \Call{FindNearestNeighbors}{R_{\text{current}}, r_{\text{forget}}}$
        \STATE
        \STATE $y_{\text{forget}} \leftarrow Y_{\text{pseudo}}[B_{\text{indices}}[idx_{\text{forget}}]]$
        \STATE $Y_{\text{heal}} \leftarrow Y_{\text{pseudo}}[\text{indices of } B_{\text{siblings}}]$
        \STATE
        \STATE $\mathcal{L}_{\text{forget}} \leftarrow \Call{CrossEntropy}{M(x_{\text{forget}}), y_{\text{forget}}}$
        \STATE $\mathcal{L}_{\text{heal}} \leftarrow \Call{CrossEntropy}{M(B_{\text{siblings}}), Y_{\text{heal}}}$
        \STATE $\mathcal{L}_{\text{unlearn}} \leftarrow \mathcal{L}_{\text{forget}} - \alpha \cdot \mathcal{L}_{\text{heal}}$
        \STATE
        \STATE $\nabla \leftarrow \Call{ComputeGradients}{-\mathcal{L}_{\text{unlearn}}}$
        \STATE $\Call{UpdateParameters}{M, \mathcal{O}, \nabla}$
    \ENDFOR
    
    \STATE \RETURN $M$
\end{algorithmic}
\end{algorithm}

In the more challenging non-targeted scenario, the knowledge to be forgotten is only implicitly defined by a dataset, $D_{\text{forget}}$, without any specific labels. The algorithm addresses this by first inferring the unwanted knowledge and then surgically removing it using a hybrid, two-phase strategy that ensures a stable and targeted unlearning process.

The first phase is a pre-computation step that establishes a fixed foundation for unlearning. The algorithm uses the initial model to generate a static set of pseudo-labels, $Y_{\text{pseudo}}$ for the entire $D_{\text{forget}}$ dataset. Concurrently, it creates a representative "fingerprint" of the unwanted knowledge by calculating the contribution map for every data point in $D_{\text{forget}}$, and averaging them into a single contribution centroid, $r_{\text{centroid}}$. These two components, the static labels and the knowledge centroid, are computed only once, preventing the optimization target from drifting and ensuring stable convergence.

The second phase is an iterative unlearning loop. In each iteration, the algorithm samples a batch $B$ and identifies the single most representative instance of the unwanted knowledge, $x_{\text{forget}}$ by selecting the data point whose contribution vector has the highest cosine similarity to the pre-computed $r_{\text{centroid}}$. This focuses the forgetting pressure on the most critical data point in each step. Subsequently, it identifies sibling knowledge, $B_{\text{siblings}}$, by finding the nearest neighbors to $x_{\text{forget}}$ within the batch's local contribution space.

The unlearning update is driven by a composite loss function that uses the static pseudo-labels for both forgetting and healing objectives. The forgetting loss, $\mathcal{L}_{\text{forget}}$, is the cross-entropy between the model's current prediction for $x_{\text{forget}}$ and its pre-computed pseudo-label $y_{\text{forget}}$. The healing loss, $\mathcal{L}_{\text{heal}}$, is the cross-entropy for the sibling data against their corresponding pseudo-labels. By performing gradient ascent on the combined loss, $\mathcal{L}_{\text{unlearn}} = \mathcal{L}_{\text{forget}} - \alpha \cdot \mathcal{L}_{\text{heal}}$, the algorithm simultaneously pushes the model to be incorrect on the most representative forgotten sample while reinforcing its original behavior on closely related points, thereby executing a precise and computationally feasible step of \textit{Neural Healing}.

\section{Discussions}\label{sec:discuss}

\textcolor{black}{\name's \textit{Neural Healing} preserves the original contribution of the nullified neurons during the unlearning by reallocating it to sibling neurons that may deliver different knowledge. This novel contribution reallocation can effectively mitigate our proposed indirect unlearning attack by maintaining the prediction accuracy of all other data, except for the forgotten data. However, the performance of this mechanism depends on the dataset quality. If the training dataset is not well balanced, the reallocation could result in either over- or under-healing.}

\textcolor{black}{In the case of under-healing, the reallocation is insufficient to fully compensate for the knowledge destruction caused by unlearning. This risk is particularly pronounced in the non-targeted method, as a biased or insufficient $D_{forget}$ can lead to an inaccurate representation of the knowledge to be removed, resulting in incomplete healing. This leaves the model vulnerable to the indirect unlearning attack, as the accuracy of non-target classes can still be significantly degraded, albeit potentially to a lesser extent than with no healing at all.
}

\textcolor{black}{Conversely, over-healing has the potential to make the model even more robust by strengthening the knowledge of classes related to the forgotten data. Achieving this positive outcome, however, is contingent on the quality of the retain dataset used for healing. If the sibling data used for reallocation is not carefully selected and representative, over-healing could inadvertently strengthen unintended biases or create new side-channel vulnerabilities. Therefore, the careful collection of the retain dataset is critical to ensuring that over-healing leads to a more robust model rather than introducing new flaws.}

\section{Evaluation} \label{sec:eval}

\subsection{Evaluation Setup}
To comprehensively evaluate the performance and robustness of our proposed \name framework, we designed a series of experiments across different domains, including image classification, zero-shot classification, and question-answering with LLMs.

\subsubsection{Models and Datasets}

\noindent \textbf{Vision and Multi-Modal Models:} 
For classification tasks, we used a range of transformer-based models: DeiT-tiny/16, ViT-base/16, CLIP-base/32, and CLIP-large/14. These models were evaluated on the CIFAR-100 and Tiny-ImageNet datasets for standard classification and zero-shot classification performance. The initial demonstration of the indirect unlearning attack (Section~\ref{sec:threat:indirect}) was conducted using the CLIP-base/32 model on the CIFAR-10 dataset.


\noindent \textbf{Large Language Models:} To assess \name's applicability in the language domain, we used Llama 3.2-1B and Llama 3.2-3B models. Their performance was evaluated on question-answering tasks from the MMLU dataset


To ensure the generalizability of our results, we performed unlearning on 10 randomly selected classes (for image classification) or subjects (for LLMs) for each model-dataset pair and averaged the results. For the unlearning process, we used a learning rate of $5 \times 10^{-7}$ which is applicable value for GA-based unlearning on relatively small-sized models.

\subsection{Unlearning Performance}

\begin{table*}[tp!]
\centering
\caption{Unlearning Performance of \name\ on Vision and Multi-Modal Classification Tasks. Performance is measured by Mean Target Accuracy (mTA), where a lower score indicates more effective forgetting, and Mean Retain Accuracy (mRA), where a score close to the baseline indicates better knowledge preservation over 10 randomly selected classes.}
\label{tab:performance:classifications}

\begin{tabular}{@{}clc|cccc|cccc@{}}
\toprule
\multirow{3}{*}{\textbf{Task}} &
  \multicolumn{1}{c}{\multirow{3}{*}{\textbf{Model}}} &
  \multicolumn{1}{c|}{\multirow{3}{*}{\textbf{\begin{tabular}[c]{@{}c@{}}Unlearning\\ Method\end{tabular}}}} &
  \multicolumn{4}{c|}{\textbf{CIFAR100}} &
  \multicolumn{4}{c}{\textbf{Tiny-ImageNet}} \\ \cmidrule(l){4-11} 
 &
  \multicolumn{1}{c}{} &
  \multicolumn{1}{c|}{} &
  \multicolumn{2}{c}{\textbf{Baseline}} &
  \multicolumn{2}{c|}{\textbf{Unlearned}} &
  \multicolumn{2}{c}{\textbf{Baseline}} &
  \multicolumn{2}{c}{\textbf{Unlearned}} \\
 &
  \multicolumn{1}{c}{} &
  \multicolumn{1}{c|}{} &
  \multicolumn{1}{c}{\textbf{Acc.}} &
  \multicolumn{1}{c}{\textbf{mTA}} &
  \multicolumn{1}{c}{\textbf{mTA}} &
  \multicolumn{1}{c|}{\textbf{mRA}} &
  \multicolumn{1}{c}{\textbf{Acc.}} &
  \multicolumn{1}{c}{\textbf{mTA}} &
  \multicolumn{1}{c}{\textbf{mTA}} &
  \multicolumn{1}{c}{\textbf{mRA}} \\ \midrule
\multirow{4}{*}{Classification} & \multirow{2}{*}{DeiT-tiny/16}  & Targeted   &  0.745&        0.73&        0.001&        0.7242&  0.6758&        0.732&        0.012&        0.6497\\
                                &                                & Non-Targeted &  0.7532&        0.692&        0.232&        0.6503&  0.6784&  0.766&  0.508&  0.6321\\ \cmidrule(l){2-11} 
                                & \multirow{2}{*}{ViT-base/16} & Targeted   &   0.8987&        0.895&        0.1589&        0.8844&  0.8731&  0.886&  0.194&  0.8657\\
                                &                                & Non-Targeted &  0.9003&   0.8817&   0.2401&        0.8979&  0.9071&  0.8299&  0.212&  0.8966\\ \midrule
\multirow{4}{*}{\begin{tabular}[c]{@{}c@{}}Zero-shot\\ Classification\end{tabular}} &
  \multirow{2}{*}{CLIP-base/32} &
  Targeted &
   \multirow{2}{*}{0.5472}&
  0.62 &
  0.0 &
  0.8057 &
   \multirow{2}{*}{0.4949}&
   0.4679&
   0.0&
   0.4473\\
                                &                                & Non-Targeted &  & 0.556 & 0.086 & 0.5284 &  & 0.422&  0.08&  0.4391\\ \cmidrule(l){2-11} 
                                & \multirow{2}{*}{CLIP-large/14} & Targeted   &   \multirow{2}{*}{0.6521}&        0.623&        0.0&        0.84302&  \multirow{2}{*}{0.6704}&  0.6839&  0.01&  0.8102\\
                                &                                & Non-Targeted &  &        0.6358&        0.1&        0.6309&  &  0.6581&  0.12&  0.6381\\ \bottomrule
\end{tabular}
\end{table*}

\begin{table*}[t!]
\centering
\caption{Unlearning Performance on on LLMs. Performance is measured by Exact Match (EM) and Pre-Quantized Exact Match (PQEM) using the MMLU dataset, averaged over 10 randomly selected target subjects for unlearning.}
\label{tab:performance:llms}
\begin{tabular}{@{}cc|cccc|cccc@{}}
\toprule
\multirow{3}{*}{\textbf{Model}} &
  \multirow{3}{*}{\textbf{\begin{tabular}[c]{@{}c@{}}Unlearning\\ Method\end{tabular}}} &
  \multicolumn{4}{c|}{\textbf{Baseline}} &
  \multicolumn{4}{c}{\textbf{Unlearned}} \\ \cmidrule(l){3-10} 
 &
   &
  \multicolumn{2}{c}{\textbf{All Subjects}} &
  \multicolumn{2}{c|}{\textbf{Target Subjects}} &
  \multicolumn{2}{c}{\textbf{Target Subjects}} &
  \multicolumn{2}{c}{\textbf{Non-Target Subjects}} \\
 &
   &
  \multicolumn{1}{c}{\textbf{EM}} &
  \multicolumn{1}{c}{\textbf{PQEM}} &
  \multicolumn{1}{c}{\textbf{mEM}} &
  \multicolumn{1}{c|}{\textbf{mPQEM}} &
  \multicolumn{1}{c}{\textbf{mEM}} &
  \multicolumn{1}{c}{\textbf{mPQEM}} &
  \multicolumn{1}{c}{\textbf{mEM}} &
  \multicolumn{1}{c}{\textbf{mPQEM}} \\ \midrule
\multirow{1}{*}{Llama 3.2-1B} &
  Targeted &
   0.2587&
   0.4218&
   0.2567&
   0.4226&
   0.2558&
   0.4183&
   0.2682&
   0.4317\\
   \midrule
\multirow{1}{*}{Llama 3.2-3B} &
  Targeted &
   0.4957&
   0.6109&
   0.5219&
   0.6369&
   0.5150&
   0.6265&
   0.4957&
   0.6117\\
  \bottomrule
\end{tabular}\vspace{-15pt}

\end{table*}

We evaluate the performance of our \name on two primary criteria: (1) its ability to effectively forget the target knowledge, measured by the mean Target Accuracy (mTA), and (2) its ability to preserve performance on all other data, measured by the mean Retain Accuracy (mRA). Lower mTA indicates more effective forgetting, while an mRA close to the baseline accuracy denotes better knowledge preservation. For each evaluation, we randomly selected 10 classes (or subjects in the case of LLMs) as unlearning targets to ensure the stability and generalizability of our results across diverse scenarios.

On standard classification tasks, as shown in Table~\ref{tab:performance:classifications}, \name demonstrates superior performance across multiple vision transformer models and datasets. For both DeiT-tiny and ViT-base models on CIFAR-100 and Tiny-ImageNet, our methods successfully reduce the accuracy on the unlearned class to near-zero levels. This is achieved with minimal impact on other classes, with the mRA remaining exceptionally close to the baseline accuracy. For instance, with ViT-base on CIFAR-100, the non-targeted method 
achieves an mRA of 0.8979, a negligible drop from the 0.9003 baseline, while effectively erasing the target class.

The benefits of our Neural Healing approach are shown in the zero-shot classification tasks using CLIP models. The targeted \name method not only achieves complete or near-complete forgetting of the target class but consistently enhances the accuracy of the retained classes. This trend holds for both model sizes. This proves that our healing process effectively reinforces retained knowledge 
by reallocating the contributions of the forgotten knowledge, leading to a more robust and accurate model.

For LLMs, we extended our evaluation to question-answering tasks using the MMLU dataset with Llama 3.2-1B and Llama 3.2-3B models. As shown in Table~\ref{tab:performance:llms}, our targeted \name method shows exceptional knowledge preservation. 
Specifically, for Llama 3.2-3B,  the performance on non-target subjects remained almost identical to the baseline. This highlights that \name successfully unlearns targeted information while maintaining the 
performance of the model on retained knowledge, thus effectively preventing indirect unlearning attacks. 

\subsection{Robustness on Imbalanced Predictions after Unlearning}

\begin{table*}[h!]
\caption{Imbalanced Prediction Ratio Analysis on CIFAR-10. Table (a) shows the baseline prediction ratio for each class before unlearning. Tables (b), (c), and (d) detail the changes in these ratios in percentage points after applying Conventional GA, our proposed Targeted Unlearning, and Non-Targeted Unlearning, respectively. Red color indicates significantly vulnerable classes, which are candidates of $C_{target}^{adv}$, under the indirect unlearning attack that showing 25\% or more accuracy change.}\label{tab:classification_ratio:roka:targeted}
\centering
\begin{tabular}{@{}l|cccccccccc@{}}
\toprule
\textbf{Label} &
  \textbf{airplane} &
  \textbf{automobile} &
  \textbf{bird} &
  \textbf{cat} &
  \textbf{deer} &
  \textbf{dog} &
  \textbf{frog} &
  \textbf{horse} &
  \textbf{ship} &
  \textbf{truck} \\ \midrule
\textbf{Prediction Ratio(\%)} &
  8.72 &
  9.65 &
  12.5 &
  8.93 &
  9.24 &
  9.59 &
  9.03 &
  12.01 &
  10.77 &
  10.56 \\ \bottomrule
\end{tabular} \\
\hspace{0.5em}
\subcaption{Baseline Prediction Ratio}

\begin{tabular}{@{}cc|cccccccccc@{}}
\toprule
 &
   &
  \multicolumn{10}{c}{\textbf{Predictions}} \\ \cmidrule(l){3-12} 
 &
   &
  airplane &
  automobile &
  bird &
  cat &
  deer &
  dog &
  frog &
  horse &
  ship &
  truck \\ \midrule
\multicolumn{1}{c|}{} &
  airplane &
  \textbf{-8.72} &
  0.15 &
  {6.76} &
  -0.35 &
  -0.16 &
  3.39 &
  -1.0 &
  -0.87 &
  0.55 &
  0.25 \\
\multicolumn{1}{c|}{} &
  automobile &
  -0.49 &
  \textbf{-9.65} &
  -0.5 &
  -1.86 &
  1.14 &
  -1.16 &
  -2.44 &
  2.35 &
  2.25 &
  {10.36} \\
\multicolumn{1}{c|}{} &
  bird &
  {\color[HTML]{333333} 5.76} &
  {11.61} &
  \textbf{-12.5} &
  -2.63 &
  -1.92 &
  5.8 &
  {\color[HTML]{FE0000} \textbf{-6.06}} &
  4.21 &
  -1.72 &
  -2.55 \\
\multicolumn{1}{c|}{} &
  cat &
  3.81 &
  -1.0 &
  1.61 &
  \textbf{-8.93} &
  1.04 &
  {\color[HTML]{FE0000} \textbf{-9.57}} &
  -3.51 &
  {9.91} &
  { 6.17} &
  0.47 \\
\multicolumn{1}{c|}{} &
  deer &
  {14.15} &
  -0.12 &
  {\color[HTML]{FE0000} -5.22} &
  {\color[HTML]{FE0000} -4.65} &
  \textbf{-9.24} &
  {\color[HTML]{FE0000} -4.92} &
  {\color[HTML]{FE0000} -5.92} &
  {17.93} &
  -1.21 &
  -0.8 \\
\multicolumn{1}{c|}{} &
  dog &
  {11.69} &
  0.05 &
  5.27 &
      {\color[HTML]{FE0000} \textbf{-7.29}} &
  0.58 &
  \textbf{-9.59} &
  -1.87 &
  1.85 &
  0.54 &
  -1.23 \\
\multicolumn{1}{c|}{} &
  frog &
  4.86 &
  5.42 &
  -0.25 &
  0.91 &
  -2.57 &
  -1.74 &
  \textbf{-8.03} &
  3.93 &
  1.21 &
  -3.74 \\
\multicolumn{1}{c|}{} &
  horse &
  1.98 &
  {6.87} &
  5.14 &
  -1.58 &
  23.2 &
  {\color[HTML]{FE0000} \textbf{-7.25}} &
  {\color[HTML]{FE0000} -4.15} &
  \textbf{-12.01} &
  -2.72 &
  -9.48 \\
\multicolumn{1}{c|}{} &
  ship &
  {49.55} &
  5.86 &
  {\color[HTML]{FE0000} \textbf{-7.96}} &
  {\color[HTML]{FE0000} -5.96} &
  {\color[HTML]{FE0000} -4.02} &
  {\color[HTML]{FE0000} \textbf{-8.35}} &
  {\color[HTML]{FE0000} -5.72} &
  {\color[HTML]{000000} -3.48} &
  \textbf{-10.77} &
  {\color[HTML]{FE0000} \textbf{-9.15}} \\
\multicolumn{1}{c|}{\multirow{-10}{*}{\textbf{Target}}} &
  truck &
  -1.31 &
  {41.88} &
  {\color[HTML]{FE0000} -4.62} &
  {\color[HTML]{000000} -2.7} &
  {\color[HTML]{000000} -2.76} &
  {\color[HTML]{FE0000} -4.75} &
  {\color[HTML]{FE0000} -5.09} &
  {\color[HTML]{000000} -3.68} &
  {\color[HTML]{FE0000} -6.41} &
  \textbf{-10.56} \\ \bottomrule
\end{tabular}\\
\hspace{0.5em}
\subcaption{Conventional Gradient Ascent}\label{tab:classification_ratio:GA}

\begin{tabular}{@{}cccccccccccc@{}}
\toprule
 &
  \multicolumn{1}{c|}{} &
  \multicolumn{10}{c}{\textbf{Predictions}} \\ \cmidrule(l){3-12} 
 &
  \multicolumn{1}{c|}{} &
  airplane &
  automobile &
  bird &
  cat &
  deer &
  dog &
  frog &
  horse &
  ship &
  truck \\ \midrule
\multicolumn{1}{c|}{} &
  \multicolumn{1}{c|}{airplane} &
  \textbf{-8.72} &
  0.54 &
  {{6.63}} &
  -0.2 &
  -0.3 &
  2.82 &
  0.49 &
  -1.54 &
  0.48 &
  -0.02 \\
\multicolumn{1}{c|}{} &
  \multicolumn{1}{c|}{automobile} &
  0.36 &
  \textbf{-9.65} &
  -1.44 &
  -1.11 &
  0.97 &
  -0.53 &
  -1.06 &
  1.13 &
  1.16 &
  {\color[HTML]{333333} 0.97} \\
\multicolumn{1}{c|}{} &
  \multicolumn{1}{c|}{bird} &
  {\color[HTML]{333333} 1.63} &
  {\color[HTML]{333333} 2.87} &
  \textbf{-12.48} &
  1.84 &
  0.53 &
  4.76 &
  {{6.19}} &
  -2.09 &
  -1.57 &
  -1.68 \\
\multicolumn{1}{c|}{} &
  \multicolumn{1}{c|}{cat} &
  1.02 &
  0.31 &
  -0.02 &
  \textbf{-8.66} &
  {\color[HTML]{333333} 1.89} &
  {\color[HTML]{333333} 0.95} &
  {\color[HTML]{333333} 3.68} &
  {\color[HTML]{333333} 1.22} &
  {\color[HTML]{333333} -0.19} &
  {\color[HTML]{333333} -0.2} \\
\multicolumn{1}{c|}{} &
  \multicolumn{1}{c|}{deer} &
  {\color[HTML]{333333} 2.81} &
  {\color[HTML]{333333} -0.22} &
  {\color[HTML]{333333} -2.52} &
  {\color[HTML]{333333} -0.38} &
  \textbf{-9.24} &
  {\color[HTML]{333333} -.075} &
  {\color[HTML]{333333} 2.04} &
  {{6.82}} &
  -0.75 &
  -0.07 \\
\multicolumn{1}{c|}{} &
  \multicolumn{1}{c|}{dog} &
  {{6.03}} &
  {\color[HTML]{333333} -0.17} &
  {\color[HTML]{333333} 0.65} &
  {\color[HTML]{333333} -3.25} &
  {\color[HTML]{333333} 2.06} &
  \textbf{-9.58} &
  {\color[HTML]{333333} 4.58} &
  0.44 &
  -0.39 &
  -0.37 \\
\multicolumn{1}{c|}{} &
  \multicolumn{1}{c|}{frog} &
  0.85 &
  0.6 &
  1.66 &
  4.72 &
  -0.99 &
  -0.09 &
  \textbf{-7.8} &
  -0.37 &
  0.13 &
  -0.69 \\
\multicolumn{1}{c|}{} &
  \multicolumn{1}{c|}{horse} &
  {\color[HTML]{333333} 0.92} &
  {\color[HTML]{333333} 1.32} &
  {\color[HTML]{333333} -2.14} &
  {\color[HTML]{333333} 2.38} &
  {\color[HTML]{333333} 9.02} &
  {\color[HTML]{333333} 1.48} &
  {\color[HTML]{333333} 2.21} &
  \textbf{-12.01} &
  -1.42 &
  -1.67 \\
\multicolumn{1}{c|}{} &
  \multicolumn{1}{c|}{ship} &
  {{8.17}} &
  {\color[HTML]{333333} 5.22} &
  {\color[HTML]{333333} -1.92} &
  {\color[HTML]{333333} -0.33} &
  {\color[HTML]{333333} -0.68} &
  {\color[HTML]{333333} 0.46} &
  {\color[HTML]{333333} 0.95} &
  {\color[HTML]{333333} -1.35} &
  \textbf{-9.86} &
  {\color[HTML]{333333} -2.68} \\
\multicolumn{1}{c|}{\multirow{-10}{*}{\textbf{Target}}} &
  \multicolumn{1}{c|}{truck} &
  0.39 &
  {{22.25}} &
  {\color[HTML]{333333} -2.9} &
  {\color[HTML]{333333} -0.38} &
  {\color[HTML]{333333} -0.62} &
  {\color[HTML]{333333} -1.09} &
  {\color[HTML]{333333} -1.87} &
  {\color[HTML]{333333} -2.36} &
  {\color[HTML]{333333} -3.62} &
  \textbf{-10.56} \\ \bottomrule
\end{tabular} \\
\hspace{0.5em}
\subcaption{Proposed Targeted Unlearning}\label{tab:classification_ratio:targeted}

\begin{tabular}{@{}cccccccccccc@{}}
\toprule
 &
  \multicolumn{1}{c|}{} &
  \multicolumn{10}{c}{\textbf{Prediction}} \\ \cmidrule(l){3-12} 
 &
  \multicolumn{1}{c|}{} &
  airplane &
  automobile &
  bird &
  cat &
  deer &
  dog &
  frog &
  horse &
  ship &
  truck \\ \midrule
\multicolumn{1}{c|}{} &
  \multicolumn{1}{c|}{airplane} &
  \textbf{-8.45} &
  -0.47 &
  10.09 &
  -0.36 &
  -0.21 &
  -0.35 &
  -1.08 & 
  -0.76 &
  -0.31 &
  -0.3 \\
\multicolumn{1}{c|}{} &
  \multicolumn{1}{c|}{automobile} &
  -0.35 &
  \textbf{-9.52} &
  -0.06 &
  -1.46 &
  0.67 &
  -0.76 &
  -.146 &
  1.48 &
  1.5 &
  {{9.69}} \\
\multicolumn{1}{c|}{} &
  \multicolumn{1}{c|}{bird} &
  -2.28 &
  1.35 &
  \textbf{-12.5} &
  -2.71 &
  -2.2 &
  1.63 &
  {{20.33}} &
  -1.07 &
  -1.64 &
  -0.91 \\
\multicolumn{1}{c|}{} &
  \multicolumn{1}{c|}{cat} &
  -0.45 &
  -0.88 &
  2.79 &
  \textbf{-8.48} &
  -0.8 &
  -0.23 &
  -0.02 &
  3.73 &
  0.23 &
  -0.22 \\
\multicolumn{1}{c|}{} &
  \multicolumn{1}{c|}{deer} &
  -0.84 &
  -0.09 &
  -3.71 &
  -2.97 &
  \textbf{-9.21} &
  -0.93 &
  {\color[HTML]{FE0000} -4.06} &
  {{21.9}} &
  -0.18 &
  -0.09 \\
\multicolumn{1}{c|}{} &
  \multicolumn{1}{c|}{dog} &
  4.77 &
  -0.35 &
  -0.19 &
  {\color[HTML]{FE0000} -4.16} &
  2.01 &
  \textbf{-9.27} &
  1.64 &
  5.710 &
  -0.44 &
  -0.31 \\
\multicolumn{1}{c|}{} &
  \multicolumn{1}{c|}{frog} &
  -0.69 &
  1.14 &
  3.25 &
  5.06 &
  -1.87 &
  0.73 &
  \textbf{-7.63} &
  0.92 &
  0.01 &
  -0.9 \\
\multicolumn{1}{c|}{} &
  \multicolumn{1}{c|}{horse} &
  1.01 &
  {{26.92}} &
  -3.84 &
  {\color[HTML]{FF0000} -5.16} &
  { 6.47} &
  {\color[HTML]{FF0000} -4.33} &
  {\color[HTML]{FF0000} -4.95} &
  \textbf{-11.54} &
  -1.69 &
  -2.89 \\
\multicolumn{1}{c|}{} &
  \multicolumn{1}{c|}{ship} &
  {\color[HTML]{FF0000} -4.3} &
  {\color[HTML]{333333} \textbf{49.08}} &
  {\color[HTML]{FF0000} \textbf{-7.4}} &
  {\color[HTML]{FF0000} -5.51} &
  {\color[HTML]{333333} -1.88} &
  {\color[HTML]{333333} -1.47} &
  {\color[HTML]{FF0000} -5.34} &
  {\color[HTML]{333333} -2.58} &
  \textbf{-10.77} &
  {\color[HTML]{FF0000} \textbf{-9.83}} \\
\multicolumn{1}{c|}{\multirow{-10}{*}{\textbf{Target}}} &
  \multicolumn{1}{c|}{truck} &
  -0.96 &
  {{15.42}} &
  -0.66 &
  0.42 &
  -0.41 &
  -0.05 &
  -1.0 &
  -0.67 &
  -1.46 &
  \textbf{-10.53} \\ \bottomrule 
\end{tabular}\\
\hspace{0.5em}
\subcaption{Proposed Non-Targeted Unlearning}\label{tab:classification_ratio:non_targeted}\vspace{-10pt}
\end{table*}

A main vulnerability of conventional unlearning methods that enables 
our indirect unlearning attack is the significant imbalances in the model's prediction behavior after unlearning. As shown in Table~\ref{tab:classification_ratio:GA}, the GA method, while effective for reducing the accuracy of the target class, causes drastic and unpredictable shifts in how other classes are classified. 
for instance, when unlearning the \texttt{ship} class, the model's prediction ratio for \texttt{airplane} increases up to 49.55\%. Similarly, unlearning \texttt{truck} causes the model to misclassify other inputs as \texttt{automobile} with a 41.88\% higher frequency. This shows that the \textit{Knowledge Destruction} signifies the prediction imbalance through increasing misclassifications of untargeted classes. 
Such an imbalanced classification is the main source of vulnerabilities leveraged by the indirect unlearning attack. 

In contrast, our proposed \name method shows significantly more balanced predictions, as shown in Table~\ref{tab:classification_ratio:targeted}. 
This stability is important for mitigating the indirect unlearning attack. By preventing drastic and imbalanced shifts in prediction accuracy, \name removes the 
side-channel that the attack relies on. An attacker can no longer reliably degrade the performance of a security-critical class $C_{target}^{adv}$ by requesting to unlearn a seemingly unrelated class $C_{unlearng}^{adv}$, because the \textit{Healing} process ensures that the model's overall knowledge structure remains balanced and robust.

\subsection{Unlearning Stability}

\begin{figure}[t]
\centering
    \subfloat[CIFAR-100 \label{fig:stability:cifar100}]{
            \includegraphics[width=0.49\linewidth]{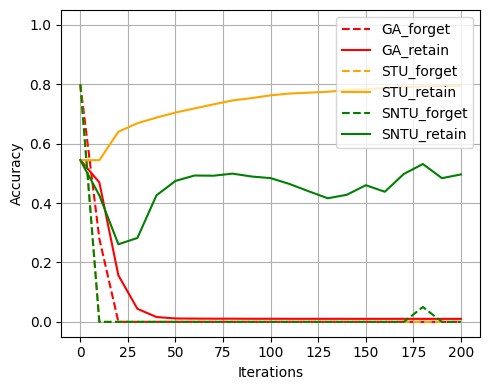}
    }
    \subfloat[Tiny-Imagenet \label{fig:stability:tiny_imagenet}]{
            \includegraphics[width=0.49\linewidth]{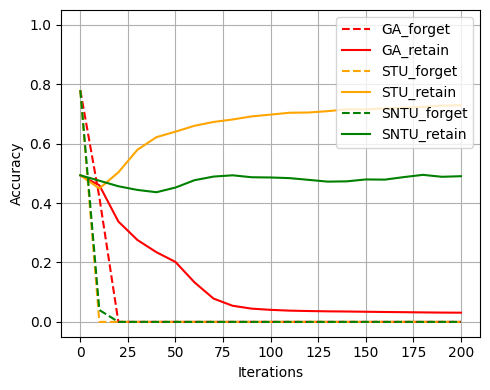}
    }   
    \caption{Unlearning Stability Comparison on CIFAR-100 and Tiny-Imagenet: The plots show the accuracy trends for the forget set (dashed lines) and the retain set (solid lines) over 200 unlearning iterations. GA indicates the results of the gradient-ascending-based unlearning, and STU and SNTU show the results of \name\ using stochastic targeted unlearning and stochastic non-targeted unlearning, respectively. }\vspace{-5pt}
    \label{fig:stability}
\end{figure}

An ideal unlearning method should not only remove target information but also maintain stable performance on the retained data throughout the process. A common failure mode, a catastrophic forgetting, occurs when the accuracy of retained data collapses as the target data is unlearned. We evaluate the stability of our proposed methods against this failure by plotting the accuracy of both the forget set and the retain set over the course of the unlearning iterations, comparing them against the conventional GA baseline.

Figure~\ref{fig:stability} shows this comparison for both the CIFAR-100 and Tiny-Imagenet datasets. The GA method (red lines) exhibits poor stability. As the accuracy of the forget set rapidly drops to zero, the accuracy of the retain set also suffers a significant and continuous decline. This demonstrates a poor trade-off, where forgetting one knowledge causes substantial collateral damage to others, confirming the problem of \textit{Knowledge Contamination}.

In contrast, our \name methods demonstrate far greater stability. The Targeted method (yellow lines) shows remarkable performance. While the forget set's accuracy quickly drops to zero, the retain set's accuracy is not only preserved but also consistently improves, eventually surpassing its initial performance. This is a clear indicator that the \textit{Neural Healing} mechanism is successfully reallocating knowledge and strengthening related concepts. The Non-Targeted method (green lines) also shows strong stability, maintaining a much steadier retain set's accuracy compared to the baseline GA, even as it forgets the target data. These results empirically validate that \name provides a more stable and robust unlearning process, effectively mitigating catastrophic forgetting of retained knowledge.



\section{Conclusion} \label{sec:con}

In this paper, we addressed the critical challenge of \textit{Knowledge Contamination}  in machine unlearning, a phenomenon that degrades model performance and introduces security vulnerabilities. We presented a novel \textit{indirect unlearning attack}, where the collateral damage from conventional unlearning methods can be exploited to compromise a model's integrity on security-critical tasks. To counter this threat, we introduced \name, a robust unlearning framework grounded in our theoretical model of \textit{Neural Knowledge System}. \name constructively reallocates the influence of forgotten data to strengthen conceptually related knowledge. Our evaluations on vision transformers and LLMs showed that \name effectively unlearns target information while robustly preserving the model's performance on retained data. This stability in performance and prediction distribution directly mitigates the threat of indirect unlearning attacks, offering a more secure and reliable alternative to conventional methods.


\bibliographystyle{plain}
\bibliography{reference}

@INPROCEEDINGS{9519428,
author={Bourtoule, Lucas and Chandrasekaran, Varun and Choquette-Choo, Christopher A. and Jia, Hengrui and Travers, Adelin and Zhang, Baiwu and Lie, David and Papernot, Nicolas},
booktitle={2021 IEEE Symposium on Security and Privacy (SP)}, 
title={Machine Unlearning}, 
year={2021},
volume={},
number={},
pages={141-159},
doi={10.1109/SP40001.2021.00019}
}

@inproceedings{10.1609/aaai.v38i11.29092,
author = {Foster, Jack and Schoepf, Stefan and Brintrup, Alexandra},
title = {Fast machine unlearning without retraining through selective synaptic dampening},
year = {2024},
isbn = {978-1-57735-887-9},
publisher = {AAAI Press},
url = {https://doi.org/10.1609/aaai.v38i11.29092},
doi = {10.1609/aaai.v38i11.29092},
booktitle = {Proceedings of the Thirty-Eighth AAAI Conference on Artificial Intelligence and Thirty-Sixth Conference on Innovative Applications of Artificial Intelligence and Fourteenth Symposium on Educational Advances in Artificial Intelligence},
articleno = {1344},
numpages = {9},
series = {AAAI'24/IAAI'24/EAAI'24}
}

@inproceedings{jang-etal-2023-knowledge,
title = "Knowledge Unlearning for Mitigating Privacy Risks in Language Models",
author = "Jang, Joel  and
Yoon, Dongkeun  and
Yang, Sohee  and
Cha, Sungmin  and
Lee, Moontae  and
Logeswaran, Lajanugen  and
Seo, Minjoon",
editor = "Rogers, Anna  and
Boyd-Graber, Jordan  and
Okazaki, Naoaki",
booktitle = "Proceedings of the 61st Annual Meeting of the Association for Computational Linguistics (Volume 1: Long Papers)",
month = jul,
year = "2023",
address = "Toronto, Canada",
publisher = "Association for Computational Linguistics",
url = "https://aclanthology.org/2023.acl-long.805/",
doi = "10.18653/v1/2023.acl-long.805",
pages = "14389--14408",
}

@article{Wang_Zeng_Guo_Wong_Gottlob_2025, title={Selective Forgetting: Advancing Machine Unlearning Techniques and Evaluation in Language Models}, volume={39}, url={https://ojs.aaai.org/index.php/AAAI/article/view/32068}, DOI={10.1609/aaai.v39i1.32068}, number={1}, journal={Proceedings of the AAAI Conference on Artificial Intelligence}, author={Wang, Lingzhi and Zeng, Xingshan and Guo, Jinsong and Wong, Kam-Fai and Gottlob, Georg}, year={2025}, month={Apr.}, pages={843-851} }

@article{bach2015pixel,
  title={On pixel-wise explanations for non-linear classifier decisions by layer-wise relevance propagation},
  author={Bach, Sebastian and Binder, Alexander and Montavon, Gr{\'e}goire and Klauschen, Frederick and M{\"u}ller, Klaus-Robert and Samek, Wojciech},
  journal={PloS one},
  volume={10},
  number={7},
  pages={e0130140},
  year={2015},
  publisher={Public Library of Science San Francisco, CA USA}
}

@article{montavon2017explaining,
  title={Explaining nonlinear classification decisions with deep taylor decomposition},
  author={Montavon, Gr{\'e}goire and Lapuschkin, Sebastian and Binder, Alexander and Samek, Wojciech and M{\"u}ller, Klaus-Robert},
  journal={Pattern recognition},
  volume={65},
  pages={211--222},
  year={2017},
  publisher={Elsevier}
}

@article{Kullback_1951,
  author = {Kullback, S. and Leibler, R. A.},
  doi = {10.1214/aoms/1177729694},
  journal = {The Annals of Mathematical Statistics},
  month = mar,
  number = 1,
  pages = {79--86},
  publisher = {Institute of Mathematical Statistics},
  title = {On Information and Sufficiency},
  url = {https://doi.org/10.1214%2Faoms%2F1177729694},
  volume = 22,
  year = 1951
}

@inproceedings{huang-2024-iclr,
title = "Unlearn and Burn: Adversarial Machine Unlearning Requests Destroy Model Accuracy",
author = "Yangsibo Huang and Daogao Liu and Lynn Chua and Badih Ghazi and Pritish Kamath and Ravi Kumar and Pasin Manurangsi and Milad Nasr and Amer Sinha and Chiyuan Zhang",
booktitle = "Proceedings of the 12th International Conference on Learning Representations (ICLR)",
month = May,
year = "2024",
address = "Vienna, Austria",
}

@inproceedings{ye-2025-security,
title = "Data Duplication: A Novel Multi-Purpose Attack Paradigm in Machine Unlearning",
author = "Dayong Ye and Tianqing Zhu and Jiayang Li and Kun Gao and Bo Liu and Leo Yu Zhang and Wanlei Zhou and Yang Zhang",
booktitle = "Proceedings of the 34th Usenix Security Symposium",
month = Aug,
year = "2025",
address = "Seattle, WA",
}

@inproceedings{song-2025-security,
title = "Refusal Is Not an Option: Unlearning Safety
Alignment of Large Language Models",
author = "Minkyoo Song and Hanna Kim and Jaehan Kim and Seungwon Shin and Sooel Son",
booktitle = "Proceedings of the 34th Usenix Security Symposium",
month = Aug,
year = "2025",
address = "Seattle, WA",
}

@inproceedings{naderloui-2025-security,
title = "Rectifying Privacy and Efficacy Measurements in Machine Unlearning: A New Inference Attack Perspective",
author = "Nima Naderloui and Shenao Yan and Binghui Wang and Jie Fu and Wendy Hui Wang and Weiran Liu and Yuan Hong",
booktitle = "Proceedings of the 34th Usenix Security Symposium",
month = Aug,
year = "2025",
address = "Seattle, WA",
}

@inproceedings{di-2023-neurips,
title = "Hidden Poison: Machine Unlearning Enables
Camouflaged Poisoning Attacks",
author = "Jimmy Z. Di and Jack Douglas and Jayadev Acharya and Gautam Kamath and Ayush Sekhari",
booktitle = "Proceedings of the 37th Conference on Neural Information Processing Systems",
month = Dec,
year = "2023",
address = "New Orleans, LA",
}

@inproceedings{zhao-2023-neurips,
title = "Static and Sequential Malicious Attacks in the Context of Selective Forgetting",
author = "Chenxu Zhao and Wei Qian and Rex Ying and Mengdi Huai",
booktitle = "Proceedings of the 37th Conference on Neural Information Processing Systems",
month = Dec,
year = "2023",
address = "New Orleans, LA",
}

@inproceedings{huang-2024-security,
title = "UBA-Inf: Unlearning Activated Backdoor Attack
with Influence-Driven Camouflage",
author = "Zirui Huang and Yunlong Mao and Sheng Zhong",
booktitle = "Proceedings of the 33rd USENIX Security Symposium",
month = Aug,
year = "2024",
address = "Philadelphia, PA",
}

@inproceedings{liul-2024-aaai,
title = "Backdoor Attacks via Machine Unlearning",
author = "Zihao Liu1 and TianhaoWang and Mengdi Huai1 and Chenglin Miao1",
booktitle = "Proceedings of the 38th AAAI Conference on Artificial Intelligence",
month = Feb,
year = "2025",
address = "Philadelphia, PA",
}

@inproceedings{pawelczyki-2025-iclr,
title = "Machine Unlearning Fails To Remove
Data Poisoning Attacks",
author = "Martin Pawelczyk1 and Jimmy Z. Di and Yiwei Lu and Ayush Sekhari and Gautam Kamath and Seth Neel",
booktitle = "Proceedings of the 13th International Conference on Learning Representations (ICLR)",
month = April,
year = "2025",
address = "Singapore EXPO",
}

@article{ccpa-2018,
  author = {Stuart L. Pardau},
  url = {https://scholarship.law.ufl.edu/jtlp/vol23/iss1/2},
  journal = {Journal of Technology Law \& Policy},
  title = {THE CALIFORNIA CONSUMER PRIVACY ACT: TOWARDS A EUROPEAN-STYLE PRIVACY REGIME IN THE UNITED STATES?},
  volume = 23,
  issue = 1,
  year = 2018
}

@article{gdpr-2011,
  author = {Jeffrey Rosen},
  url = {https://www.stanfordlawreview.org/online/privacy-paradox-the-right-to-be-forgotten/},
  journal = {Stanford Law Review},
  title = {The right to be forgotten},
  year = 2011
}

@misc{touvron2021training,
      title={Training data-efficient image transformers \& distillation through attention}, 
      author={Hugo Touvron and Matthieu Cord and Matthijs Douze and Francisco Massa and Alexandre Sablayrolles and Hervé Jégou},
      year={2021},
      eprint={2012.12877},
      archivePrefix={arXiv},
      primaryClass={cs.CV}
}

@misc{dosovitskiy2021imageworth16x16words,
      title={An Image is Worth 16x16 Words: Transformers for Image Recognition at Scale}, 
      author={Alexey Dosovitskiy and Lucas Beyer and Alexander Kolesnikov and Dirk Weissenborn and Xiaohua Zhai and Thomas Unterthiner and Mostafa Dehghani and Matthias Minderer and Georg Heigold and Sylvain Gelly and Jakob Uszkoreit and Neil Houlsby},
      year={2021},
      eprint={2010.11929},
      archivePrefix={arXiv},
      primaryClass={cs.CV},
      url={https://arxiv.org/abs/2010.11929}, 
}

@misc{radford2021learningtransferablevisualmodels,
      title={Learning Transferable Visual Models From Natural Language Supervision}, 
      author={Alec Radford and Jong Wook Kim and Chris Hallacy and Aditya Ramesh and Gabriel Goh and Sandhini Agarwal and Girish Sastry and Amanda Askell and Pamela Mishkin and Jack Clark and Gretchen Krueger and Ilya Sutskever},
      year={2021},
      eprint={2103.00020},
      archivePrefix={arXiv},
      primaryClass={cs.CV},
      url={https://arxiv.org/abs/2103.00020}, 
}

@misc{grattafiori2024llama3herdmodels,
      title={The Llama 3 Herd of Models}, 
      author={Aaron Grattafiori and Abhimanyu Dubey and Abhinav Jauhri and Abhinav Pandey and Abhishek Kadian and Ahmad Al-Dahle and Aiesha Letman and Akhil Mathur and Alan Schelten and Alex Vaughan and Amy Yang and Angela Fan and et al.},
      year={2024},
      eprint={2407.21783},
      archivePrefix={arXiv},
      primaryClass={cs.AI},
      url={https://arxiv.org/abs/2407.21783}, 
}

@inproceedings{Le2015TinyIV,
  title={Tiny ImageNet Visual Recognition Challenge},
  author={Ya Le and Xuan S. Yang},
  year={2015},
  url={https://api.semanticscholar.org/CorpusID:16664790}
}

@misc{hendrycks2021measuringmassivemultitasklanguage,
      title={Measuring Massive Multitask Language Understanding}, 
      author={Dan Hendrycks and Collin Burns and Steven Basart and Andy Zou and Mantas Mazeika and Dawn Song and Jacob Steinhardt},
      year={2021},
      eprint={2009.03300},
      archivePrefix={arXiv},
      primaryClass={cs.CY},
      url={https://arxiv.org/abs/2009.03300}, 
}

@TECHREPORT{Krizhevsky09learningmultiple,
    author = {Alex Krizhevsky},
    title = {Learning multiple layers of features from tiny images},
    institution = {},
    year = {2009}
}

@inproceedings{wang2024attributebyunlearning,
  title={Data Attribution for Text-to-Image Models by Unlearning Synthesized Images},
  author={Wang, Sheng-Yu and Hertzmann, Aaron and Efros, Alexei A and Zhu, Jun-Yan and Zhang, Richard},
  booktitle={NeurIPS},
  year = {2024},
}

\end{document}